%% file: sample-base.tex
\documentclass[sigconf]{acmart}
\usepackage{enumitem}
\usepackage{multirow}
\usepackage{subfigure}
\usepackage{algorithm}
\usepackage{algorithmic}
\usepackage{tabularx}
\usepackage{array}
\usepackage{makecell}
\usepackage{graphicx}

\usepackage{xcolor}
\usepackage{amsmath}
\usepackage{amsthm}

\newcolumntype{C}{>{\centering\arraybackslash}X}
\newcolumntype{P}[1]{>{\centering\arraybackslash}p{#1}}

\AtBeginDocument{%
  \providecommand\BibTeX{{%
    \normalfont B\kern-0.5em{\scshape i\kern-0.25em b}\kern-0.8em\TeX}}}

\setcopyright{acmcopyright}
\copyrightyear{2018}
\acmYear{2018}
\acmDOI{XXXXXXX.XXXXXXX}

\acmConference[Conference acronym 'XX]{Make sure to enter the correct
  conference title from your rights confirmation emai}{June 03--05,
  2018}{Woodstock, NY}
\acmBooktitle{Woodstock '18: ACM Symposium on Neural Gaze Detection,
 June 03--05, 2018, Woodstock, NY} 
\acmPrice{15.00}
\acmISBN{978-1-4503-XXXX-X/18/06}

\begin{document}

\title{Self-Explainable Graph Neural Networks for Link Prediction}

\author{Huaisheng Zhu}
\affiliation{The Pennsylvania State University   \country{USA}
}
\email{hvz5312@psu.edu}

\author{Dongsheng Luo}
\affiliation{Florida International University   \country{USA}
}
\email{dluo@fiu.edu }

\author{Xianfeng Tang}
\affiliation{Amazon  \country{USA}
}
\email{xianft@amazon.com}

\author{Junjie Xu}
\affiliation{The Pennsylvania State University   \country{USA}
}
\email{junjiexu@psu.edu}

\author{Hui Liu}
\affiliation{Michigan State University   \country{USA}}
\email{liuhui7@msu.edu}

\author{Suhang Wang}
\affiliation{The Pennsylvania State University   \country{USA}}
\email{szw494@psu.edu}

\renewcommand{\shortauthors}{Trovato and Tobin, et al.}

\begin{abstract}
Graph Neural Networks (GNNs) have achieved state-of-the-art performance for link prediction. However, GNNs suffer from poor interpretability, which limits their adoptions in critical scenarios that require knowing why certain links are predicted. Despite various methods proposed for the explainability of GNNs, most of them are post-hoc explainers developed for explaining node classification. Directly adopting existing post-hoc explainers for explaining link prediction is sub-optimal because: (i) post-hoc explainers usually adopt another strategy or model to explain a target model, which could misinterpret the target model; and (ii) GNN explainers for node classification identify crucial subgraphs around each node for the explanation; while for link prediction, one needs to explain the prediction for each pair of nodes based on graph structure and node attributes. Therefore, in this paper, we study a novel problem of self-explainable GNNs for link prediction, which can simultaneously give accurate predictions and explanations. Concretely, we propose a new framework and it can find various $K$ important neighbors of one node to learn pair-specific representations for links from this node to other nodes. These  $K$ different neighbors represent important characteristics of the node and model various factors for links from it. Thus, $K$ neighbors can provide explanations for the existence of links. Experiments on both synthetic and real-world datasets verify the effectiveness of the proposed framework for link prediction and explanation.
\end{abstract}





\maketitle

\input{introduction_v2}

\input{related_work.tex}

\section{Problem Definition}
We use $\mathcal{G}=(\mathcal{V}, \mathcal{E}, \mathbf{X})$ to denote an attributed graph, where $\mathcal{V}=\left\{v_{1}, \ldots, v_{N}\right\}$ is the set of $N$ nodes, $\mathcal{E}$ is the set of edges and $\mathbf{X}$ is the attribute matrix for nodes in $\mathcal{G}$. The $i$-th row of $\mathbf{X}$, i.e., $\mathbf{x}_i\in \mathbb{R}^{1\times d_0}$, is the $d_0$ dimensional features of node $v_i$. $\mathbf{A} \in \mathbb{R}^{N \times N}$ is the adjacency matrix. $A_{ij}=1$ if node $v_i$ and node $v_j$ are connected; otherwise $A_{ij}=0$. $\mathcal{N}_i$ represents the neighborhood set of $v_i$. The goal of link prediction is to determine whether there exists an edge $e_{ij}$ between two given nodes $\{v_i, v_j\}$. It can be formulated as a classification problem on a set of node pairs $\mathcal{E}_U$ given observed edges $\mathcal{E}_L$ and node attributes, where $e_{ij}= 1$ represents a link between node $v_i$ and $v_j$, and  $e_{ij}= 0$ means no link between $v_i$ and $v_j$. %where $e_{ij} \in \mathcal{E}_U$ or $\mathcal{E}_L$ and the corresponding element in $\mathbf{A}$ is $A_{ij}=0$, or  1. 
%Most of existing GNNs focus on link prediction of unobserved edges in $\mathcal{E}_U$. 
Due to the great node representation learning ability, GNNs are usually adopted as an encoder: $f: \mathcal{V} \rightarrow \mathbb{R}^d$ to map a node $v_i$ to a $d$-dimensional vector $\mathbf{h}_{i}$ for link prediction. $f$ should preserve the similarity between nodes based on the observed edge set $\mathcal{E}_L$ \cite{zhou2020graph}, and give large probability (large ($f(v_i)^Tf(v_j)$)) for $e_{ij}=1$ but small probability (small ($f(v_i)^Tf(v_j)$)) for $e_{ij}=0$. However, GNN usually lacks interpretability on why they give such predictions \cite{kipf2016variational, zhang2018link}. There are few attempts of explainers on the node classification task~\cite{dai2021towards,ying2019gnnexplainer, luo2020parameterized, zhang2022protgnn}, while the work on interpretable link prediction based on GNNs is rather limited \cite{ wang2021modeling}. Thus, it's crucial to develop interpretable GNNs for link prediction.
% \dongsheng{Since a more detailed definition is provided at the end, here , we can briefly describe the problem at a high level}

% We aim to develop interpretable link prediction for Graph Neural Networks (GNNs). 
%  

As mentioned in the introduction, generally, the preferences of a node are reflected in its neighbors. For node pairs, only some of the neighbors with common features are important for their link prediction. Therefore, to learn pair-specific representation for each node pair $(v_i, v_j)$, we propose to find the top $K$ neighbors of node $v_i$ which are similar to $v_j$. Specifically, for a node pair $(v_i, v_j)$, 
we can learn pair-specific representations $\mathbf{h}_i$ and $\mathbf{h}_j$ by aggregating selected top $K$ neighbors. Thus, $\mathbf{h}_i^T\mathbf{h}_j$ will be large when neighbors of $v_i$ are similar to $v_j$ and dissimilar neighbors will result in lower $\mathbf{h}_i^T\mathbf{h}_j$. Due to the undirected property of graphs, it also holds true for node $v_j$. \textit{Explanations of link prediction can be}: \textit{(i)} ``for a node pair $v_i$ and $v_j$ with a link, take node $v_i$ as an example, $v_i$'s neighbors $v_c \in \mathcal{N}_i$ have higher similarity score  with regard to $v_j$ and $v_j$ in the graph. This explanation also holds for node $v_j$''; \textit{(ii)} ``for a node pair $v_i$ and $v_j$ without a link, neighbors of $v_i$ are dissimilar to $v_j$ and also neighbors of $v_j$ are dissimilar to $v_i$''. 
% and high order or highly non-linear structural similarity score
%  The representation of a node pair $(v_i, v_j)$ is determined by aggregating their neighbors information so we want to find $K$ neighbor nodes of $v_i$ and $v_j$ separately which are relevant to the existence of links between them. Specifically, for node $v_i$, we propose to find $K$ neighbor nodes of $v_i$ which are similar or near to node $v_j$. Thus, $f(v_i)^Tf(v_j)$ will be large when neighbors of $v_i$ are similar to $v_j$ and dissimilar neighbors will result in lower value of $f(v_i)^Tf(v_j)$. Due to the undirected properties of graphs, it also holds true for node $v_j$. Explanations of link prediction can be: (i) ``for a node pair $v_i$ and $v_j$ with a link, they have common neighbors. Then, take node $v_i$ as an example, $v_i$'s neighbors $v_c \in \mathcal{N}_i$ have higher feature similarity score and high order or highly non-linear structural similarity score with regard to $v_j$ or near to $v_j$ in the graph. This explanation also holds for node $v_j$''; (ii) ``for a node pair $v_i$ and $v_j$ without a link, neighbors of $v_i$ are dissimilar to $v_j$ and also neighbors of $v_j$ are dissimilar to $v_i$''. 
 
With the notations above, the problem is formally defined as: 
\textit{Given an attributed graph $\mathcal{G}=(\mathcal{V}, \mathcal{E}_L, \mathbf{X})$ with observed edge set $\mathcal{E}_L$ and unobserved edge set $\mathcal{E}_U$, learn an interpretable link predictor $g_{\theta}: \mathcal{V} \times \mathcal{V} \rightarrow \{True, False\}$  which can accurately predict links in $\mathcal{E}_U$ and simultaneously generate explanation by identifying two sets of important $K$ neighbors for  link between each node pair $(v_i, v_j)$. %denoted by $\mathcal{N}^r_i$ and $\mathcal{N}^r_j$, where $v_c \in \mathcal{N}^r_i$ are the identified top $K$ nodes similar to $v_j$ and it also holds true for $v_c \in \mathcal{N}^r_j$ due to the undirected property of the graph. 
}

\section{Methodology}

\begin{figure*}[t!] 
\centering 
\vskip -1em
\includegraphics[width=0.92\textwidth]{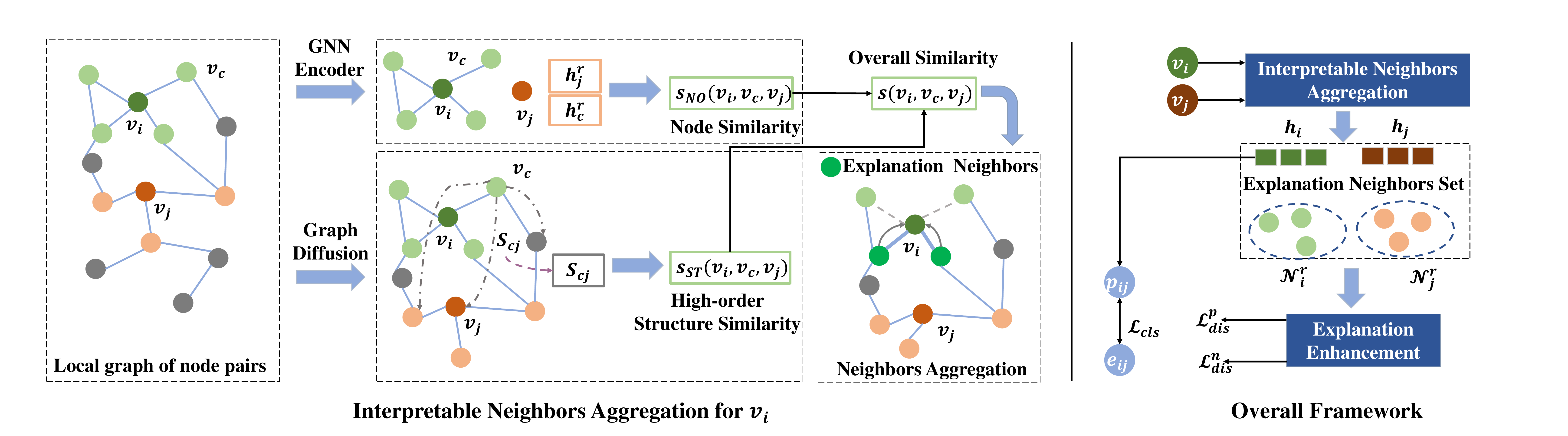}
\vskip -1.6em
\caption{An overview of the proposed ILP-GNN.} 
\label{model} 
\vskip -1em
\end{figure*}

In this section, we introduce the details of the proposed framework ILP-GNN. The basic idea of ILP-GNN is: for each node pair $(v_i, v_j)$, it identifies $K$ neighbors of node $v_i$ and $v_j$, respectively, aiming to capture the common interests of the two nodes. Then, it aggregates these $K$ neighbors' information to obtain their pair-specific representation vectors and calculate the similarity based on their representations. Meanwhile, the $K$ most relevant neighbors of $v_i$ and $v_j$ provide the explanation on why there is or isn't a link between this node pair. There are mainly two challenges: (i) how to obtain the $K$ most relevant neighbors of $v_i$ and $v_j$ for the prediction of links between them; (ii) how to simultaneously give accurate predictions and correct corresponding explanations.  To address these challenges, for each node pair $(v_i,v_j)$, ILP-GNN explicitly models both node and high-order structure similarity between $v_i$ and neighbors of $v_j$ to identify the $K$ important nodes of $v_j$ for explainable link prediction. Similarly, it explicitly models both node and high-order structure similarity between $v_j$ and neighbors of $v_i$ to identify $K$ important nodes of $v_i$.

An illustration of the proposed framework is shown in Figure~\ref{model}. It is mainly composed of an Interpretable Neighbors Aggregation module and an Explanation Enhancement module. With the Interpretable Neighbors Aggregation, for each node pair $(v_i, v_j)$, the $K$ most important neighbors, which represent factors for the existence of links, are found based on both node and high-order structure information. Then, the prediction of links between nodes can be given based on the identified $K$ neighbors. Finally, the Explanation Enhancement module is designed to further benefit the accurate explanation generation, and also encourage the model to improve link prediction via aggregating $K$ important neighbors.
\subsection{Interpretable Neighbors Aggregation}
For GNN-based link prediction on a pair of nodes $(v_i, v_j)$, we firstly aggregate their neighbors' information to obtain their representation vectors $(\mathbf{h}_i, \mathbf{h}_j)$. Then, the similarity will be calculated between $(\mathbf{h}_i, \mathbf{h}_j)$ via $\mathbf{h}_i^T\mathbf{h}_j$ to indicate whether there is a link between them and pair-specific representation is required to predict links. 

However, links are generated due to multiple factors. 
For different links from $v_i$, it's necessary for us to learn pair-specific representations for $v_i$ to predict links between them. In graph data, the neighborhood of $v_i$ can represent important characteristics but not all neighbors have relevant factors w.r.t different connected nodes of $v_i$. Based on this observation, ILP-GNN selects  $K$ neighbors of $v_i$ which are similar to $v_j$ to learn pair-specific representation of $v_i$. Similarly, through the same way, we learn a pair-specific representation of $v_j$ with $K$ interpretable neighbors. ILP-GNN relies on interpretable $K$ neighbors that reveal the common interest of $v_i$ and $v_j$ for link prediction and explanation. We need to design a similarity measurement to measure the similarity between neighbors of node $v_i$ and another connected or unconnected node $v_j$. Unlike i.i.d data, which only needs to measure the similarity from the feature perspective, for graph-structure data, both node attributes and graph structures of nodes contain crucial information for link prediction. In the following part, for a node pair $(v_i, v_j)$, we use $v_i$ as an example to demonstrate the process of finding $K$ interpretable neighbors. And we will do the same operations for node $v_j$. 

\subsubsection{High-order Structure Similarity}
In the graph, for a pair of nodes $(v_i, v_j)$, the distance between neighbors of $v_i$ and $v_j$ can be used to measure their similarity. For instance, one-hop neighbors of the node $v_i$ may have higher similarity scores with $v_i$, while high-order neighbors will have lower similarity scores. And the one-hop relation is represented as the adjacency matrix $\mathbf{A}$ and high-order relations can be represented as $\mathbf{A}^2, \mathbf{A}^3$, etc. Since original edge relations are often sparse and missed in real-world graphs, only using one-hop neighbors to measure the distance between nodes may result in unreliable similarity, i.e., the similarity between $v_i$ and $v_j$ equals $A_{ij}$. Therefore, it's necessary to model both one-hop and high-order relations to measure the similarity between nodes based on graph structure.
To measure this high-order similarity, we propose to use a Graph Diffusion matrix which  calculates the closeness of nodes in the graph structure by repeatedly passing the weighting coefficients to the neighboring nodes and represents the high-order similarity between nodes based on graph structure\cite{klicpera2019diffusion}:
\begin{equation}
    \mathbf{S }= \sum_{k=0}^{\infty} \theta_k \mathbf{T}^k,
    \label{eq:1}
\end{equation}
where $\mathbf{T}$ represents the random walk transition matrix as $\mathbf{T}=\mathbf{A} \mathbf{D}^{-1}$, and the degree matrix $\mathbf{D}$ is the diagonal matrix of node degrees, i.e. $D_{i i}=\sum_{j=1}^N A_{i j}$. In this paper, we utilize a popular example of graph diffusion, Personalized PageRank (PPR) \cite{page1999pagerank}. PPR chooses $\theta_k^{\mathrm{PPR}}=\gamma(1-\gamma)^k$ with teleport probability $\gamma \in(0,1)$. $\gamma$ is set as 0.05 in the experiment. Then, we normalize this diffusion matrix via $\mathbf{\tilde{S}} = \mathbf{D}_{S}^{-1 / 2} \mathbf{S} \mathbf{D}_{S}^{-1 / 2}$ to convert the similarity score to $(0, 1)$. After getting this normalized diffusion matrix, for each $v_c\in \mathcal{N}_i$, the structure importance weight of $v_c$ to $v_i$ for the link of $(v_i,v_j)$ is:
 \begin{equation}
     s_{\text{ST}}(v_i, v_c, v_j) =  \tilde{S}_{cj},
     \label{eq:2}
 \end{equation}
where $\tilde{S}_{cj}$ represents high-order structure similarity between $v_i$'s neighbor $v_c$ and $v_j$ based on graph structure.
 \subsubsection{Node Similarity}
 Generally, node similarity on the feature level can be used to measure how similar neighbors of one node are to another connected or unconnected node, which can be used to find $K$ interpretable nodes relevant to the existence of links. Since node features are often noisy and sparse, directly utilizing the raw feature may result in noisy similarity. A straightforward way is to model the node's local relationships via a GNN such as GCN~\cite{kipf2016semi} and GAT~\cite{velivckovic2017graph} to learn the node embedding. However, GNN models will implicitly model graph structure information and reduce the interpretability of node features. Therefore, we first use a Multilayer Perceptron (MLP) to encode node features and then update node embedding via graph structure information. This can be written as:
\begin{equation}
    \mathbf{H}^{m}=\text{MLP}(\mathbf{X}), \quad \mathbf{H}^{r}=\sigma(\tilde{\mathbf{A}} [\mathbf{H}^{m} \| \mathbf{X}] \mathbf{W})+\mathbf{H}^{m},
    \label{eq:3}
\end{equation}
where $\|$ is the concatenation operation of two vectors. The
concatenation records local information with attribute information and facilitates training by providing skip connections. $\mathbf{H}^r$ represents the learned embedding matrix for all nodes. Then,  the similarity between neighborhoods of node $v_i$ and node $v_j$ based on node attributes and their local topology can be calculated as:
\begin{equation}
    s_{\text{NO}}(v_i, v_c, v_j) = \text{sigmoid}\big((\mathbf{h}^r_{j})^T \mathbf{h}^r_{c}\big), ~\forall~ v_c \in \mathcal{N}_i
    \label{eq:4}
\end{equation}
where $\text{sigmoid}(\cdot)$ is the sigmoid function to convert the similarity between two vectors into $(0, 1)$, and $s_{\text{NO}}(v_i, v_c, v_j)$ represents the weight $v_c$ contributes to the prediction of links between $v_i$ and $v_j$ based on node similarity. 

Finally, with node and high-order structure similarity, the importance score of $v_i$ with $v_c \in \mathbf{N}_i$  on $(v_i,v_j)$ is:
\begin{equation}
    s(v_i, v_{c}, v_j) = \alpha \cdot s_{\text{ST}}(v_i, v_c, v_j) + (1-\alpha) \cdot s_{\text{NO}}(v_i, v_{c}, v_j),
    \label{eq:5}
\end{equation}
where $\alpha$ is the hyperparameter to control the contributions of node and high-order structure similarity. 

After getting weights for neighbors of node $v_i$, we will find $K$ neighbors based on these weights. We will do the same thing for neighbors of $v_j$. In the link prediction task, common neighbors between $v_i$ and $v_j$ may be highly relevant to links between pairs of nodes~\cite{lu2009similarity}. Thus, we first select common neighbors to a new neighborhood set $\mathcal{N}^r_i$. If the number of common neighbors is larger than $K$, we can utilize top $K$ common neighbors based on weight scores. Then,  we can select the top relevant neighborhoods based on the weight scores of original neighbors of node $v_i$ to the neighborhood set $\mathcal{N}^r_i$, which makes the size of $\mathcal{N}^r_i$ equal $K$.  Note that nodes in the $\mathcal{N}^r_i$ only appear one time. If the size of $\mathcal{N}^r_i$ is smaller than $K$, we will use all of its neighborhoods with different weights. We normalize the weight scores $b_{i c}$ on this set for comparable scores:
\begin{equation}
    b_{i c}=\frac{\exp \left(s \left(v_{i}, v_{c}, v_{j}\right) \right)}{\sum_{v_c \in \mathcal{N}_{i}^r} \exp \left(s \left(v_{i}, v_{c}, v_{j}\right) \right)}.
    \label{eq:6}
\end{equation}
Finally, node $v_i$'s representation vectors can be obtained as:
\begin{equation}
    \mathbf{h}_{i}=\mathbf{h}^r_i + \beta \sum_{v_c \in \mathcal{N}_{i}^r} b_{i c} \mathbf{h}^r_c,
    \label{eq:7}
\end{equation}
where $\beta$ is used to control the influence of neighborhoods on the final representation of nodes. Also, we can use the same way to obtain the representation vector $\mathbf{h}_j$ of node $v_j$. As $\mathbf{h}_{i}$ and $\mathbf{h}_{j}$ captures important neighbors for link prediction, then the link probability $p_{ij}$ for $(v_i,v_j)$ can be calculated as:
\begin{equation}
    p_{ij} = \text{sigmoid} (\mathbf{h}_i^T\mathbf{h}_j).
    \label{eq:8}
\end{equation}

\subsection{Explanation Enhancement}
To guarantee the selected neighbors of two nodes $(v_i,v_j)$ are highly relevant to whether there is a link between them, we propose self-supervision to enhance the explanation.  
First, for each linked node pair $(v_i, v_j)$ with $e_{ij}=1$, to make sure that the selected neighbors by ILP-GNN have a positive effect on the final prediction, we randomly select $K$ neighbors of node $v_i$ from $\mathcal{N}_i \backslash \mathcal{N}_i^r$, where $\mathcal{N}_i \backslash \mathcal{N}_i^r$ represents the set of $v_i$'s neighbors excluding those selected by ILP-GNN.

We do the same operation for node $v_j$. Let the obtained two random neighbor sets be $\{\mathcal{N}^{\text{rand}}_i, \mathcal{N}^{\text{rand}}_j\}$ for nodes $v_i$ and $v_j$, respectively. Note that there are nodes whose $|\mathcal{N}_i| - |\mathcal{N}_i^r|$ is smaller than $K$ and we will not perform the following objective function. Then we will use $\{\mathcal{N}^{\text{rand}}_i, \mathcal{N}^{\text{rand}}_j\}$ to learn node representation of $v_i$ and $v_j$ to predict link between $(v_i,v_j)$. This predicted link probability should be smaller than that of using $\{\mathcal{N}^{r}_i, \mathcal{N}^{r}_j\}$ as we expect $\{\mathcal{N}^{r}_i, \mathcal{N}^{r}_j\}$  to be more effective than $\{\mathcal{N}^{\text{rand}}_i, \mathcal{N}^{\text{rand}}_j\}$ . Specifically, the node representation of $v_i$  by aggregating randomly selected neighbors $\mathcal{N}_{i}^{\text{rand}}$ can be obtained as:
\begin{equation}
    \mathbf{h}^{\text{rand}}_{i}=\mathbf{h}^{r}_i  + \beta \sum_{v_c \in \mathcal{N}_{i}^{\text{rand}}} b_{i c}^{\text{rand}} \mathbf{h}^r_c,
\end{equation}
where $b_{i c}^{\text{rand}}$ is the weighted scores normalized on $\mathcal{N}_{i}^{\text{rand}}$ using Eq.(\ref{eq:6}). $\mathbf{h}_i^{\text{rand}}$ is $v_i$'s representation by aggregating randomly selected neighbors of $v_i$.  Then we can calculate the predicted link existence probability $p^{\text{rand}}_{ij}$ for $v_i$ and $v_j$ using $\mathbf{h}_i^{\text{rand}}$ and $\mathbf{h}_j^{\text{rand}}$ as:
\begin{equation}
    p^{\text{rand}}_{ij} = \text{sigmoid} ((\mathbf{h}_i^{\text{rand}})^T\mathbf{h}_j^{\text{rand}}).
\end{equation}
Intuitively, we would expect $\mathbf{h}_i$ and $\mathbf{h}_j$ using $\{\mathcal{N}^{r}_i, \mathcal{N}^{r}_j\}$ to be more effective for link prediction than $\mathbf{h}_i^{\text{rand}}$ and $\mathbf{h}_j^{\text{rand}}$ using randomly selected neighbors. In other word, $p_{ij}$ should be larger than that $p_{ij}^{\text{rand}}$ by a margin $\delta$ with $0<\delta<1$; otherwise, we penalize our model.  
This can be mathematically written as:
\begin{equation}
    \mathcal{L}^p_{\text{dis}} = \sum_{e_{ij}\in \mathcal{E}_L, e_{ij}=1} \max(0, p_{ij}^{\text{rand}} + \delta - p_{ij}),
    \label{eq:10}
\end{equation}

Second, for each node pair $(v_i, v_j)$ without a link, we hope that our model gives a lower predicted probability for $\mathbf{h}_i$ and $\mathbf{h}_j$ learned from $\{\mathcal{N}^{r}_i, \mathcal{N}^{r}_j\}$. This predicted probability is also determined by the similarity between $\mathbf{h}_i$ and $\mathbf{h}_j$ in Eq.(\ref{eq:8}). In other words, if  nodes $v_i$ and $v_j$ have lower similarity scores with nodes in $\mathcal{N}^r_j$ and $\mathcal{N}^r_j$ respectively, the similarity of $\mathbf{h}_i$ and $\mathbf{h}_j$ will be small and the model will give a lower probability for the link of $(v_i, v_j)$.  Therefore, the selected neighbors set $\mathcal{N}^r_i$ of node $v_i$ should be assigned lower similarity scores w.r.t node $v_j$. It also holds true for node $v_j$. To achieve this purpose, we randomly sample unlinked pairs $e_{ij} =0$ which have the same number as the number of linked pairs $e_{ij} =1$ in $\mathcal{E}_L$. The set of randomly selected unlinked pairs can be denoted as $\mathcal{E}_N$. The similarity scores can be minimized through the following loss function: 
\begin{equation}
\label{eq:12}
    \mathcal{L}_{\text{dis}}^n = \sum_{e_{ij} \in \mathcal{E}_N} \Big( \sum_{v_c \in \mathcal{N}_i^r} s (v_i, v_c, v_j)^2 + \sum_{v_c \in \mathcal{N}_j^r} s(v_j, v_{c}, v_i)^2 \Big).
\end{equation}

\subsection{Overall Objective Function}
Link prediction can be treated as a binary classification problem.  Since the majority of node pairs are unconnected, most of the elements in adjacency matrix are $0$. To avoid the the missing links dominating the loss function, following~\cite{zhang2018link}, we adopt negative sampling to alleviate this issue. We treat each linked pair in $\mathcal{E}_L$ as positive samples. For each positive sample, we randomly sample one unlinked pair as the negative sample. Then, we treat link prediction as a binary classification problem to predict positive and negative samples.

A cross-entropy loss is adopted for this binary classification problem and the loss can be written as:
\begin{equation}
    \mathcal{L}_{\text{cls}} = \sum_{e_{ij}\in \mathcal{E}_L}- \log p_{ij}+ \sum_{e_{ij}\in \mathcal{E}_N} - \log \left(1-p_{ij}\right),
\end{equation}

where $p_{ij}$ is the predicted probability for the node pair $(v_i, v_j)$ in the Eq.(\ref{eq:8}) and $\mathcal{E}_N$ is the set of negative samples in Eq.(\ref{eq:12}). The final loss function of ILP-GNN is given as: 

\begin{equation}
    \min_{\Theta} \mathcal{L} = \mathcal{L}_{\text{cls}} + \lambda (\mathcal{L}_{\text{dis}}^p + \mathcal{L}_{\text{dis}}^n),
    \label{eq:13}
\end{equation}
where $\lambda$ controls the balance between classification loss and the loss to enhance the explanation for links between node pairs, and $\Theta$ is the set of  learnable parameters for our proposed ILP-GNN.

\subsection{Training Algorithm and Time Complexity}
\begin{algorithm}[t]
\caption{ Training Algorithm of ILP-GNN.} 
\begin{algorithmic}[1]
\REQUIRE
$\mathcal{G}=(\mathcal{V},\mathcal{E}_L, \mathbf{X})$, $K$, $\lambda$, $\alpha$, $\delta$
\ENSURE GNN model $g_{\theta}$ with explanation for link prediction.
\STATE Randomly initialize the model parameters $\Theta$.
\STATE Calculate high-order distance via Eq.(\ref{eq:1}).
\REPEAT
\STATE For each node pair $(v_i, v_j)$, assign weights $s_{\text{ST}}(v_i, v_c, v_j)$ to neighbors of $v_i$ by high-order structure similarity in Eq.(\ref{eq:2}).
\STATE Learn node feature representation by Eq.(\ref{eq:3}) and assign weights to neighbors of $v_i$ by node similarity in Eq.(\ref{eq:4}).
\STATE Do the same operation on $v_j$ and aggregate top $K$ neighbors of $v_i$ and $v_j$ with two kinds of weights in Eq.(\ref{eq:7}). 
\STATE Calculate the probability $p_{ij}$ of a link between two nodes.
\STATE Randomly choose neighbors except from  top $K$ neighbors to obtain $p_{ij}^{\text{rand}}$ and calculate $\mathcal{L}^p_{\text{dis}}$ in Eq.(\ref{eq:10})%maximize difference $\mathcal{L}^p_{dis}$ between $p_{ij}^{rand}$ and $p_{ij}$ via Eq.(\ref{eq:10}) for positive samples
\STATE Calculate $\mathcal{L}_{\text{dis}}^n$  using negative samples
\STATE Update $\Theta$ by minimizing the overall loss function in Eq.(\ref{eq:13})
\UNTIL convergence
\RETURN $g_{\theta}$
\end{algorithmic}
\label{alg:1}
\end{algorithm}
\subsubsection{Training Algorithm} The training algorithm of ILP-GNN is given in Algorithm~\ref{alg:1}. We utilize $e_{ij}=1$ as positive samples and $e_{ij}=0$ as negative samples. ILP-GNN assigns weights to neighbors of nodes based on both node and high-order structure similarity. The top $K$ neighbors with high weights are aggregated to learn representation for self-explainable link predictions. Specifically, in line one, we initialize the parameters of the model with Xavier initialization~\cite{glorot2010understanding}. Then, we calculate high-order similarity by Eq.(\ref{eq:1}). In lines 4 to line 6, for each node pair $(v_i, v_j)$ with or without links, we assign weights to neighbors of $v_i$ and $v_j$ by learned node and high-order structure similarity. Then, ILP-GNN aggregates top $K$ neighbors with high weights to obtain pair-specific representation vectors $\mathbf{h}_i, \mathbf{h}_j$. Also, these top $K$ neighbors, which represent common interests between node pairs, can be treated as the explanation for the existence of links. In line 7, the probability of a link is calculated based on node representation vectors in Eq.(\ref{eq:8}). To guarantee the quality of explanation, in lines 8 and 9, for positive samples, $\mathcal{L}^p_{\text{dis}}$ is proposed to make the predicted probability from ILP-GNN larger than the probability predicted from representation vectors with randomly selected neighbors. For negative samples, $\mathcal{L}^n_{\text{dis}}$ is applied to make weights of neighbors small which represents neighbors of one node are dissimilar to another node. In this case, the model will give low predicted probabilities for the existence of links. Finally, the model is optimized on the total loss function Eq.(\ref{eq:13}).

\subsubsection{Time Complexity Analysis}
The main time complexity of our model comes from calculating the node similarity and high-order structure similarity together with our proposed loss function. For high-order structure similarity, the time complexity of Personalized PageRank (PPR) is denoted as $\mathcal{O}(k|\mathcal{E}_L|)$, where $k$ is the number of iterations in Eq.(\ref{eq:1}). Also, high-order structure similarity can be pre-computed which will not influence the training process of the model. For node similarity,  the time complexity for links $e_{ij} \in \mathcal{E} $ is  $\mathcal{O}(\sum_{e_{ij} \in \mathcal{E}} |\mathcal{N}_i| |\mathcal{N}_j|d)$, where $d$ is the embedding dimension and $|\mathcal{N}_i|$ is the number of neighbors for $v_i$. The time complexity of the proposed loss function is $\mathcal{O}(K|\mathcal{E}_L|d)$.  Therefore, the overall time complexity for the training phase in each iteration is  $\mathcal{O}(\sum_{e_{ij} \in \mathcal{E}_L} |\mathcal{N}_i| |\mathcal{N}_j|d + K|\mathcal{E}_L|d)$. The time complexity of the testing phase is $\mathcal{O}(\sum_{e_{ij} \in \mathcal{E}_U} |\mathcal{N}_i| |\mathcal{N}_j|d)$. A detailed training time comparison is given in Appendix~\ref{app:time}. 

\input{experiment.tex}

\section{Conclusion}
 In this paper, we study a novel problem of self-explainable GNNs for link prediction by exploring $K$ important neighbors for links. We propose a novel framework, which designs an interpretable aggregation module for finding $K$ neighbors relevant to factors of links and simultaneously uses these neighbors to give predictions and explanations. Also, a novel loss function is proposed to enhance the generation of explanations and also the performance of the link prediction task. Extensive experiments on real-world and synthetic datasets verify the effectiveness of the proposed ILP-GNN for explainable link prediction. An ablation study and parameter sensitivity analysis are also conducted to understand the contribution of the proposed modules and sensitivity to the hyperparameters. There are several interesting directions that need further investigation. One direction is to extend ILP-GNN for dynamic network link prediction. Another direction is to design more efficient and learnable approaches to explore high-order similarity in graphs.

% \section{Appendices}

% \begin{acks}
% \end{acks}

\bibliographystyle{ACM-Reference-Format}
\bibliography{acmart}

\appendix
\clearpage

\input{appendix}
\end{document}

%% file: introduction_v2.tex
\section{Introduction}

Graphs are pervasive in a wide spectrum of applications such as social networks~\cite{qu2021imgagn}, recommendation system~\cite{fan2019graph} and knowledge graphs~\cite{nickel2015review}. 
As real-world graphs are often partially observed, link prediction, which aims to predict missing links, has been recognized as a fundamental task. 
For example, link prediction can be used to recommend new friends on social media~\cite{adamic2003friends}, predict protein interactions~\cite{qi2006evaluation} or reconstruct knowledge graphs~\cite{nickel2015review}. Existing link prediction methods can be generally split into two categories, i.e., heuristic-based approaches~\cite{lu2009similarity, newman2001clustering, adamic2003friends, katz1953new} and representation learning based approaches~\cite{acar2009link, kipf2016variational, zhang2018link, pan2022neural}. 
Recently, due to the great ability of representation learning on graphs, Graph Neural Networks (GNNs) have achieved state-of-the-art performance for link prediction~\cite{kipf2016variational, zhang2018link}. 
Generally, GNNs iteratively update a node's representation by aggregating its neighbors' information. The learned representations capture both node attributes and local topology information, which facilitate link prediction.  

Despite their success in link prediction, GNNs are not interpretable, which hinders their adoption in various domains. For instance, in financial transaction networks, providing explainable transaction links among customers can help transaction platforms learn the reasons for transaction behaviors to improve customers' experience, and gain the credibility of platforms~\cite{maree2020towards}. 
Though various methods have been proposed for the explainability of GNNs~\cite{luo2020parameterized, ying2019gnnexplainer, schlichtkrull2020interpreting, yuan2020xgnn, yuan2021explainability}, they mainly focus on post-hoc GNN explainers for node classification. Directly adopting existing post-hoc GNN explainers to explain link prediction is sub-optimal because: (i) post-hoc explainers usually adopt another strategy or model to explain a target model, which could misinterpret the target model~\cite{dai2021towards}; and (ii) GNN explainers for node classification usually identify crucial subgraphs of each node for the explanation; while for link prediction, one needs to explain the prediction for each pair of nodes based on their features and local structures. 
Thus,  in this paper, we aim to develop a self-explainable GNN for link prediction, which can simultaneously give predictions and explanations.

\begin{figure}[t] 
\centering 
\includegraphics[width=0.45\textwidth]{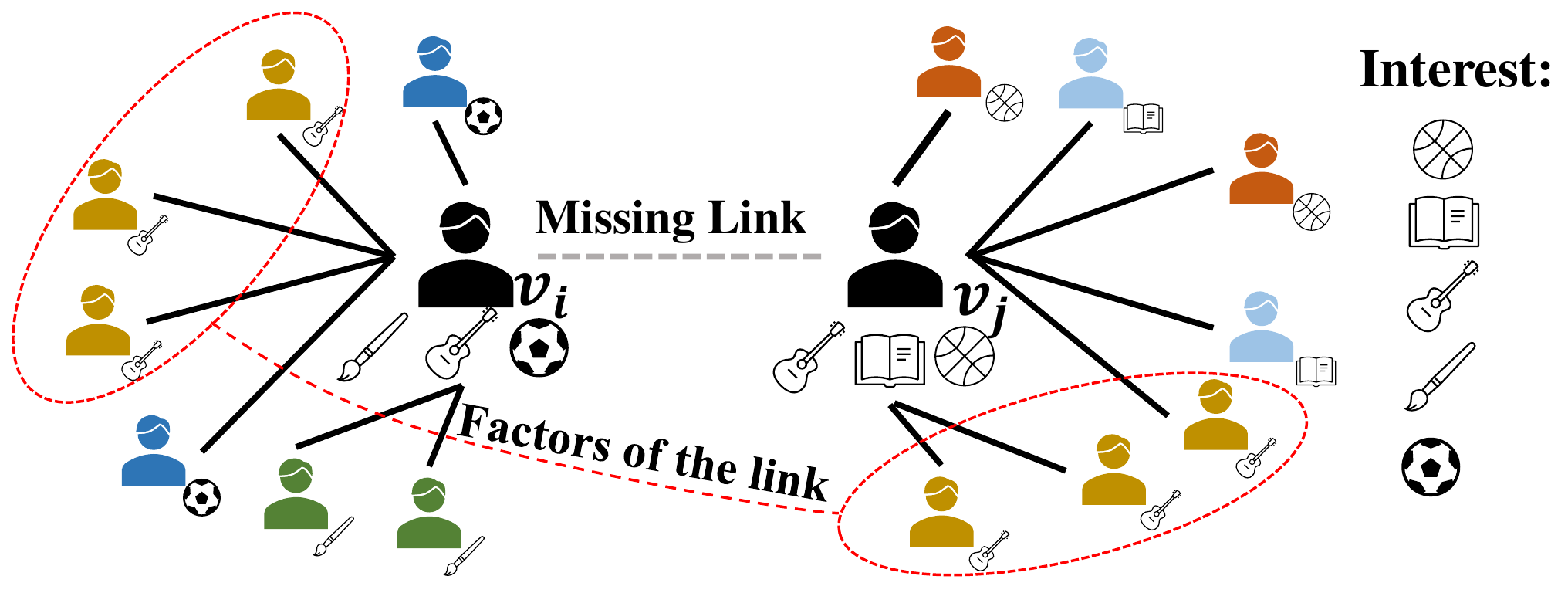} 
\vskip -1.6em
\caption{Example of factors for the link between $v_i$ and $v_j$.} 
\label{illustration}
\vskip -1em
\end{figure}

In real-world graphs, nodes are linked due to various factors, and capturing the common factor between a pair of nodes paves us a way for explainable link prediction. For example, as shown in Figure~\ref{illustration}, user $v_i$ has broad interests in football, arts, and music. $v_i$ links to diverse neighbors based on one or two common interests with them. 
For users $v_i$ and $v_j$ in the figure, they share common interests in music and are likely to be linked. However, it is difficult to predict the link simply based on their node feature similarity as most of their features/interests are different. Also, as $v_i$ and $v_j$ have diverse neighbors, directly applying a GNN by aggregating all neighbors' information to learn representations of $v_i$ and $v_j$ followed by the similarity of node representation will result in a low predicted link probability. Thus, how to effectively capture the common factor/preferences for a pair of nodes remains a question.

Generally, the preferences of users are reflected in their neighbors. For example, both $v_i$ and $v_j$ have many neighbors interested in music. By identifying and aggregating these neighbors, we can better learn \textit{pair-specific} node representations that reflect common interests to predict the link between them.  Pair-specific means that the representation of $v_i$ for $(v_i,v_j)$ should reflect the common interest with $v_j$, and for $(v_i,v_k)$ should reflect the comment interest with $v_k$. 
Based on this intuitive principle, for each node pair $(v_i, v_j)$, one way for self-explainable link prediction is to find dominating $K$ neighbors of $v_j$ that share the highest $K$ similarities with $v_i$, which reflects the common interest of $(v_i, v_j)$ and vice versa. Then, the explanation for whether there is a link for $(v_i, v_j)$ can be: \textit{(i)} ``We suggest $v_j$ to $v_i$ because these $K$ neighbors of $v_i$ have common features with $v_j$.  $v_i$ will most likely establish a new link with $v_j$ for $v_j$ is similar to current neighbors of $v_i$; and \textit{(ii)} ``We suggest no link predicted  for $(v_i, v_j)$ because even top $K$ similar neighbors of $v_i$ w.r.t $v_j$ don't share common feature with $v_j$, which represent $v_j$ is different from neighbors of $v_i$. Therefore, it's hard for $v_i$ to add $v_j$ (different from $v_i$'s current neighbors) as a new neighbor of it." %Simialr explanation also holds true for the node $v_j$. 
Though promising, the work on exploring $K$ relevant neighbors for self-explainable GNNs on link prediction is rather limited.

Therefore, in this paper, we investigate a novel problem of self-explainable GNN for link prediction. Specifically, for a pair of nodes $(v_i,v_j)$, we want to identify $K$ most important neighbors of $v_j$ and $v_i$, respectively, for link prediction. %Similarly, this  self-explainable GNN should also explicitly identifies $K$ important neighbors of $v_j$ which are similar to node $v_i$. 
In essence, there are two main challenges: (\textit{i}) 
How to take both graph structure and node attributes into consideration when measuring node similarity for identifying important neighbors for link prediction?

and (\textit{ii}) how to give both accurate predictions and correct corresponding explanations given that we lack the supervision of groundtruth explanations? 
In an attempt to solve the challenges, we propose a novel framework named Interpretable Link Prediction based on Graph Neural Networks (ILP-GNN). For each node pair $(v_i, v_j)$, ILP-GNN adopts a novel mechanism that can explicitly evaluate the node similarity and high-order structure similarity to find $K$ interpretable neighbors of $v_i$ similar to $v_j$. These neighbors can represent common interests or features between $v_i$ and $v_j$. For high-order structure similarity, graph diffusion is utilized to calculate the closeness of nodes by modeling their local and high-order neighbors' information~\cite{klicpera2019diffusion}. Then, these $K$ neighbors are aggregated to learn pair-specific representation, which represents common features of node pairs and various factors of links for different node pairs. Furthermore, since explanation based on selecting $K$ important neighbors should preserve factors resulting in the existence of links,  we propose a novel loss function to encourage explanation of our model and improve the performance of link prediction. The main contributions are:
\begin{itemize}[leftmargin=*]
\item We study a novel problem of self-explainable GNN for link prediction by finding $K$ neighbors which are relevant to the links. 
\item We develop a novel framework ILP-GNN, which adopts an interpretable neighbor aggregation method, and a novel loss function to identify $K$ relevant neighbors for explainable link prediction;
\item We conduct extensive experiments  to demonstrate the effectiveness of our model on both predictions and explanations. We also construct a synthetic dataset that can quantitatively evaluate the link prediction explanation.
\end{itemize}

%% file: related_work.tex
\section{Related Works}
\noindent\textbf{Graph Neural Networks.} Graph Neural Networks (GNNs) have shown great ability in representation learning on graphs. Generally, GNNs can be split into two categories, i.e., spectral-based~\cite{bruna2013spectral,kipf2016semi, tang2019chebnet, he2021bernnet} and spatial-based~\cite{velivckovic2017graph, hamilton2017inductive, gao2018large, xiao2021learning, zhao2022exploring}. Spectral-based approaches are defined according to graph signal processing. \citeauthor{bruna2013spectral}~\citep{bruna2013spectral} firstly proposes convolution operation to graph data from the spectral domain. Then, a first-order approximation is utilized to simplify the graph convolution via GCN~\cite{kipf2016semi}. Spatial-based GNN models aggregate information of the neighbor nodes~\cite{hamilton2017inductive, velivckovic2017graph}. For example, the attention mechanism is utilized in Graph Attention Network (GAT) to update the node representation from the neighbors with different weights~\cite{velivckovic2017graph}. Moreover, various spatial
methods are proposed for further improvements~\cite{chen2018fastgcn, zhao2022exploring, xiao2021learning}. For instance, DisGNN models latent factors of edges to facilitate node classification~\cite{zhao2022exploring}.

\noindent\textbf{Link Prediction.} Link prediction has been widely applied in social networks~\cite{adamic2003friends} and knowledge graph~\cite{nickel2015review}. Existing methods for link prediction can be generally split into two categories, i.e., heuristic-based approaches~\cite{lu2009similarity, newman2001clustering, adamic2003friends, katz1953new} and representation learning based approaches~\cite{acar2009link, kipf2016variational, zhang2018link, pan2022neural}. Heuristics-based approaches mainly compute the pairwise similarity scores based on graph structure or node properties~\cite{lu2011link}. For example, the common-neighbor index (CN) scores a pair of nodes by the number of shared neighbors~\cite{newman2001clustering};  CN and some methods, i.e.  Adamic Adar (AA)~\cite{adamic2003friends}, are calculated from up to two-hop neighbors of the target nodes. Other approaches also explore high-order neighbors, including Katz, rooted PageRank (PR)~\cite{brin1998anatomy} and SimRank (SR)~\cite{jeh2002simrank}. But these heuristic-based methods make strong assumptions and can't be generalized to different graph data. Representation learning based approaches firstly learn the representation of nodes and then apply dot product between two node representations to predict the likelihood of the link between two nodes.  GNNs are applied to learn node-level representations that capture both the topology structure together with node feature information and achieve state-of-the-art performance on link prediction~\cite{kipf2016variational, zhang2018link, pan2022neural} in recent years. For example, VGAE~\cite{kipf2016variational} adopts GNNs to encode graph structure with features into node representations followed by a simple inner product decoder to get the link prediction result. SEAL~\cite{zhang2018link} extracts subgraphs to predict links between nodes. However, these approaches are also not interpretable and will limit their ability in applications which may require why the model predicts links between nodes. 

\noindent\textbf{Explainability of Graph Neural Networks.} To address the problem of lacking interpretability in GNNs, extensive works have been proposed~\cite{huang2022graphlime, ying2019gnnexplainer, luo2020parameterized}.  For example, GNNExplainer~\cite{ying2019gnnexplainer} learns soft masks for edges and node features to find the crucial subgraphs and features to explain the predictions.  PGExplainer~\cite{luo2020parameterized} generates the edge masks with parameterized explainer to find the significant subgraphs. However, previous methods are post-hoc explanations that learn an explainer to explain the outputs of a trained GNN with fixed parameters. Post-hoc explanations might result in unstable interpretation as generated explanations are not directly from the model. To fill this gap, self-explainable GNNs are proposed to make predictions and explanations simultaneously~\cite{dai2021towards, zhang2022protgnn}. For example, SE-GNN finds interpretable labeled neighbors which have the same labels as target nodes~\cite{dai2021towards}. But self-explainable GNN models on the link prediction task are rather limited. CONPI~\cite{wang2021modeling} models similarity between neighbors set of node pairs to determine the probability of the existence of links and provide explanations based on similar neighbors. Their explanations are based on local topology similarity which ignores high-order graph structure information. Other relevant papers are about explainable link prediction for knowledge graphs~\cite{rossi2022explaining, zhang2019interaction}. They are designed for knowledge graphs and their explanations are based on reasoning paths or (head, relation, tail) data format. Therefore, it's hard for them to be generalized to all link prediction tasks.

Our work is inherently different from the aforementioned explainable GNN methods: (i) we focus on learning a self-explainable GNN on link prediction which can simultaneously give predictions and explanations while most of the previous methods are designed for node classification; (ii) we study a novel self-explainable method to find factors which determine links between nodes by considering both node and high-order structure information.

%% file: experiment.tex
\section{Experiment}
In this section, we conduct experiments on real-world and synthetic datasets to verify the effectiveness of ILP-GNN. In particular, we aim to answer the following research questions: (\textbf{RQ1}) Can our proposed method provide accurate predictions for link prediction? 
(\textbf{RQ2}) Can ILP-GNN learn reasonable explanation for the existence of links?
(\textbf{RQ3}) How do two similarity measurement methods of our ILP-GNN contribute to the link prediction performance?

\begin{table*}[t]
    \centering
    \small
    %\tiny
    \caption{Link Prediction performance (AUC(\%) $\pm$ Std.) on all graphs.} \label{main_results}
    \vskip -1.5em
    \resizebox{2\columnwidth}{!}{
    \begin{tabular}{lccccccccc}
    \toprule
    \textbf{Method} & AA & CN & VGAE & GCN & GAT & SEAL & CONPI-Pair & WP & ILP-GNN \\
    \midrule
        Cora & 75.80 $\pm$ 0.15 & 75.57 $\pm$ 0.13 & 91.18 $\pm$ 0.41 & 91.57 $\pm$ 0.69 & 91.87 $\pm$ 0.93 & 92.21 $\pm$ 1.23 & 89.69 $\pm$ 0.32 & 92.42 $\pm$ 1.1 & \textbf{93.21 $\pm$ 1.14}  \\
        Citeseer & 69.80 $\pm$ 0.22 & 69.70 $\pm$ 0.23 & 91.42 $\pm$ 0.96 & 92.51 $\pm$ 1.00 & 93.57 $\pm$ 0.64 & 90.52 $\pm$ 1.29 & 87.47 $\pm$ 0.10 & 91.37 $\pm$ 0.98 & \textbf{95.23 $\pm$ 1.33} \\
        Photo & 96.59 $\pm$ 0.22 & 96.21 $\pm$ 0.04 & 97.03 $\pm$ 0.15 & 97.08 $\pm$ 0.13 & 96.47 $\pm$ 0.19 & 98.04 $\pm$ 0.70 & 96.45 $\pm$ 0.42 & 98.12 $\pm$ 0.14 & \textbf{98.23 $\pm$ 0.04}\\
        Ogbn-arxiv & 82.41 $\pm$ 0.02 & 82.43 $\pm$ 0.01 & 95.05 $\pm$ 0.07 & 95.27 $\pm$ 0.06 & 94.79 $\pm$ 0.03 & 95.30 $\pm$ 0.04 & 94.22 $\pm$ 0.08 & 95.33 $\pm$ 0.02 & \textbf{95.42 $\pm$ 0.03} \\
        Ogbn-collab & 58.09 $\pm$ 0.09 & 57.88 $\pm$ 0.02 & 96.27 $\pm$ 0.07 & 96.73 $\pm$ 0.02 & 96.64 $\pm$ 0.05 & 93.57 $\pm$ 0.04 & 92.31 $\pm$ 0.03 & 96.81 $\pm$ 0.02 & \textbf{97.17 $\pm$ 0.01}\\
        \bottomrule
    \end{tabular}}
    %\vskip -1em
\end{table*}

\subsection{Datasets} \label{sec:datasets}
We conduct experiments on four publicly available real-world datasets and their details are shown below:
\begin{itemize}[leftmargin=*]
    \item \textbf{Cora and Citeseer}~\cite{kipf2016semi}: These two datasets are citation networks where nodes are papers and edges are their citation relations.  Each node has a sparse bag-of-words feature vector.
    \item \textbf{Photo}~\cite{shchur2018pitfalls}: This dataset is a subgraph of the Amazon co-purchase graph~\cite{mcauley2015image}, where nodes are products and two frequently purchased products are connected via an edge. Each node has a bag-of-word feature vector of reviews. 
    \item \textbf{Ogbn-arxiv}~\cite{hu2020open}: It is a citation network between all Computer Science arXiv papers indexed by MAG \cite{wang2020microsoft}. Nodes in this dataset represent papers and edges indicate one paper citing another one. Each paper has a 128-dimensional feature vector obtained by averaging the embeddings of words in its title and abstract.
    \item \textbf{Ogbl-collab}~\cite{hu2020open}: This dataset is a subset of the collaboration network between authors indexed by MAG \cite{wang2020microsoft}, where nodes are authors and edges indicate the collaboration between authors. All nodes have 128-dimensional feature vectors by averaging the embeddings of words in papers published by authors.
\end{itemize}
%\suhang{you need to explain why we need a synthetic dataset before you introduce it}

\textbf{Synthetic-Data}: Since the publicly available datasets don't have groundtruth explanations, to quantitatively evaluate the explanation of the proposed method, we also construct synthetic datasets that have groundtruth explanation for links. We firstly generate a set of $N$ nodes, $\{v_1, v_2, ..., v_N\}$, with their feature information. Then, for the generation of links between the node pair $(v_i, v_j)$, the links are determined by $B$ neighbors of $v_i$ and these $B$ neighbors share both high structure and node similarity with $v_j$. Similarly, due to the undirected property of the graph, we also consider $B$ neighbors of $v_j$ similar to $v_i$ for the generation of links. And these $B$ neighbors of $v_i$ and $B$ neighbors of $v_j$ can be treated as groundtruth explanation for this link. We fix $N$ as 1000 and generate three datasets with different number of edges for experiments. The detailed construction process of the synthetic datasets is given in Appendix~\ref{app:data}.

The statistics of these datasets are summarized in Table~\ref{tab:datasets} in Appendix~\ref{app:data}. Following \cite{kipf2016variational}, for Cora, Citeseer and Photo, we randomly split the edges of each dataset into 85\%/5\%/10\% as train/val/test. The random split is conducted 5 times and average performance will be reported. For Ogbn-arxiv, we randomly split the edges into 60\%/10\%/30\% as train/val/test. For Ogbl-collab, follow~\cite{hu2020open}, we split the edges according to time, where the collaboration edges until 2017 are used as training edges, those in 2018 are used as validation edges, and those in 2019 are test edges. For Synthetic datasets, we randomly select $40\%$ edges from edges set with groundtruth explanation as testing set and $10\%$ as validation set. The remaining edges are used as training samples. We maintained consistency across datasets by following VAGE's split proportion and randomly selecting node pairs not in the training set as negative samples following~\cite{kipf2016variational}. Note that this approach differs from WP and SEAL, which selected pairs not present in any of the training, validation, or test sets. The number of negative samples is  equal to the number of positive samples. Positive and negative samples are then combined as our training, validation and testing sets~\cite{kipf2016variational}. 

\subsection{Experimental Setup}
\textbf{Baselines.} We  compare the proposed framework with representative and state-of-the-art methods for link prediction, which include:%with classical heuristics, GNN-based representation learning methods and GNN-based link prediction methods, which includes:
\begin{itemize}[leftmargin=*]
    \item \textbf{CN} \cite{newman2001clustering}: Common-neighbor index counts the number of common neighbors to predict the link for a pair of nodes.
    \item \textbf{AA} \cite{adamic2003friends}: Adamic–Adar index is a second-order traditional heuristic method.  It assumes that a shared neighbor with the large degree is less significant to the measure of a link.
    \item \textbf{VGAE} \cite{kipf2016variational}:  VGAE is a generative model for graph representation. We use a GCN as the encoder where the second layer has two channels for mean and deviations to sample the latent embeddings and a link reconstruction module as the decoder.
    \item \textbf{GCN} \cite{kipf2016semi}: GCN is one of the most popular spectral GNN models based on graph Laplacian, which has shown great performance for node classification. To adopt it for link prediction, we treat it as the encoders in the Graph Autoencoder manner.
    %\suhang{how do you use it for link prediction?}
    \item \textbf{GAT} \cite{velivckovic2017graph}: Instead of treating each neighbor node equally, GAT utilizes an attention mechanism to assign different weights to nodes in the neighborhood during the aggregation step.  
    \item \textbf{SEAL} \cite{zhang2018link}: SEAL is a link prediction method that extracts local subgraphs of node pairs and learns link prediction heuristics from them automatically.
    \item \textbf{CONPI} \cite{wang2021modeling}: CONPI is an interpretable model to compare the similarity between neighbors sets of two nodes for link prediction. It has two variants and we report the best results of them.
    \item \textbf{WP} \cite{pan2022neural}: WalkPooling (WP)  jointly encodes node representations and graph topology into learned topological features. Then, these features are used to enhance representation of extracted subgraphs which are relevant to links of node pairs .
\end{itemize}

As the work on explainable GNN for link prediction is rather limited, we also adapt a popular post-hoc explanation model GNNExplainer for post-hoc link prediction.
\begin{itemize}[leftmargin=*]
    \item \textbf{GNNExplainer}~\cite{ying2019gnnexplainer}: GNNExplainer takes a trained GNN and the predictions as input to obtain post-hoc explanations. A soft edge mask is learned for each instance to identify the crucial subgraph. We adopt GNNExplainer to find crucial neighbors of node pairs with links for link prediction.
\end{itemize}
\noindent\textbf{Configurations.} All experiments are conducted on  a 64-bit machine with Nvidia GPU (NVIDIA RTX A6000, 1410MHz , 48 GB memory). For a fair comparison, we utilize a two-layer graph neural network for all methods, and the hidden dimension is set as 128. The learning rate is initialized to 0.001. Besides, all models are trained until converging, with the maximum training epoch as 1000. The implementations of all baselines are based on Pytorch Geometric or their original code. The hyperparameters of all methods are tuned on the validation set. In particular, for the proposed framework, we select $K$ from 1 to 6 and vary $\lambda$ as $\{0.1, 0.3, 0.5, 0.7\}$. The $\alpha$ which balances the node similarity and high-order structure similarity is fixed as 0.3 for all datasets. The margin $\delta$ in Eq.(\ref{eq:10}) is set as 0.5.
%\suhang{talk a little bit about the grid search for our method}

\subsection{Performance on Link Prediction}
In this subsection, we compare the performance of the proposed ILP-GNN with baselines for link prediction on real-world graphs introduced in Sec.~\ref{sec:datasets}, which aims to answer \textbf{RQ1}. Each experiment is conducted 5 times for all datasets and the average link prediction AUC scores with standard deviations are reported in Table~\ref{main_results}. From the table, we make the following observations:
\begin{itemize}[leftmargin=*]
    \item  Our method outperforms VGAE, GCN and GAT on various real-world
datasets. This is because for each node pair $(v_i, v_j)$, our model can select $K$ neighbors of $v_i$ and $v_j$ separately, which contain common characteristics of $(v_i, v_j)$ and are highly relevant to the factors for the existence of the link. By aggregation these $K$ significant neighbors, our model can learn pair-specific representations of $v_i$ and $v_j$ for the link and predict it accurately.
% \suhang{huaisheng, can you make it clear use notations, readers can easily get lost when you say one node, another connected or unconnected node? Why do we keep mention connected, unconnected node?} 
\item ILP-GNN can outperform SEAL and its variant WP which extracts subgraphs to learn link prediction heuristics. The reason is that our model can implicitly find local neighbors which can also explore link prediction heuristics like common neighbors and high-order structure information. 
\item Though CONPI also compares neighbors of node pairs and finds relevant neighbors for link prediction, our model can outperform CONPI, which shows the effectiveness of our method in selecting neighbors for link prediction and our loss function can guide the model to obtain link prediction relevant neighbors. 
\end{itemize} 

\subsection{Explanation Quality}

\begin{figure}[t]
\centering
\subfigure[Cora]{
\begin{minipage}[t]{0.45\linewidth}
\centering
\vskip -2pt 
\includegraphics[width=\textwidth]{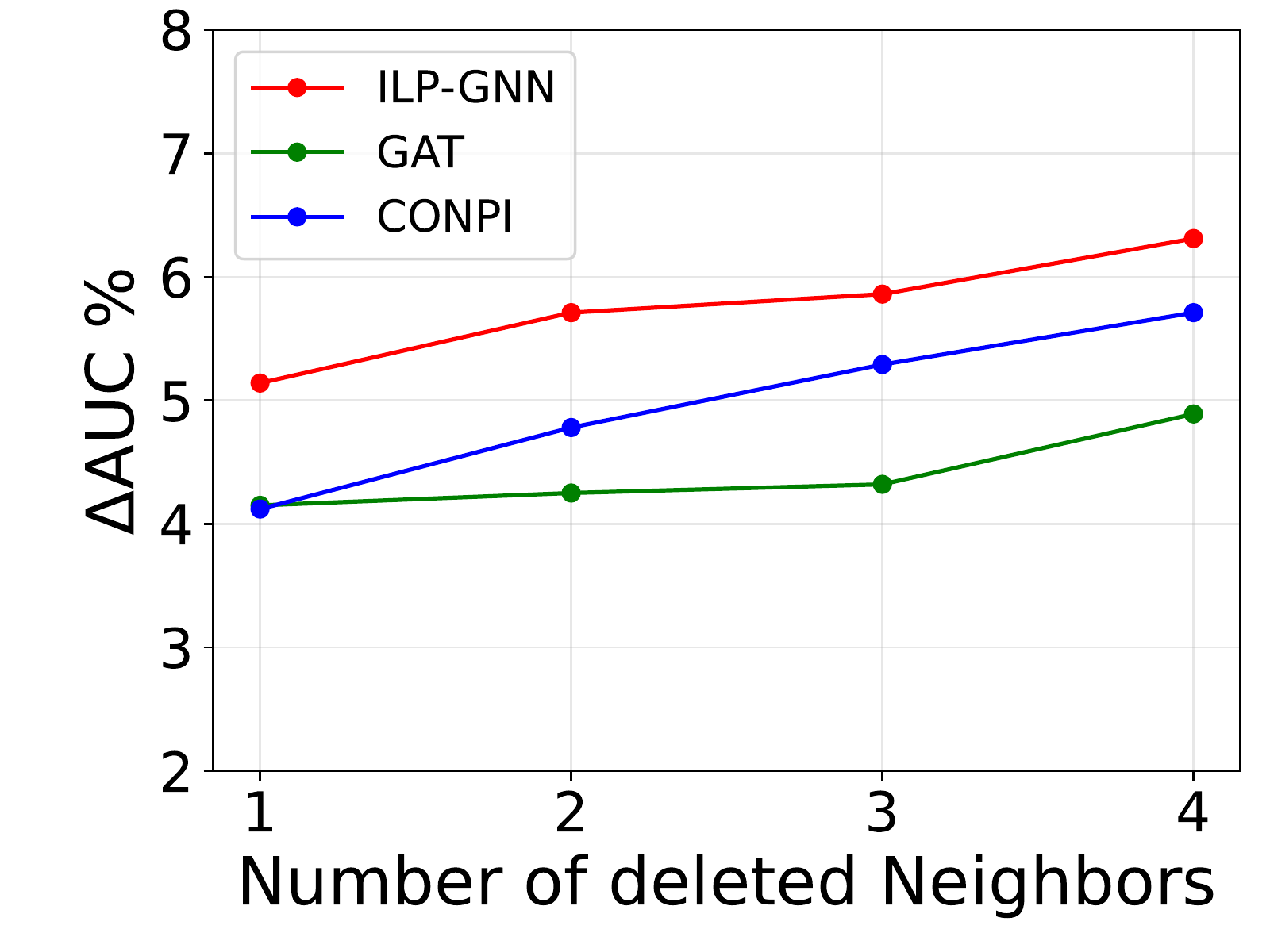}
\end{minipage}%
}~~%
\subfigure[Photo]{
\begin{minipage}[t]{0.46\linewidth}
\centering
\vskip -2pt 
\includegraphics[width=\textwidth]{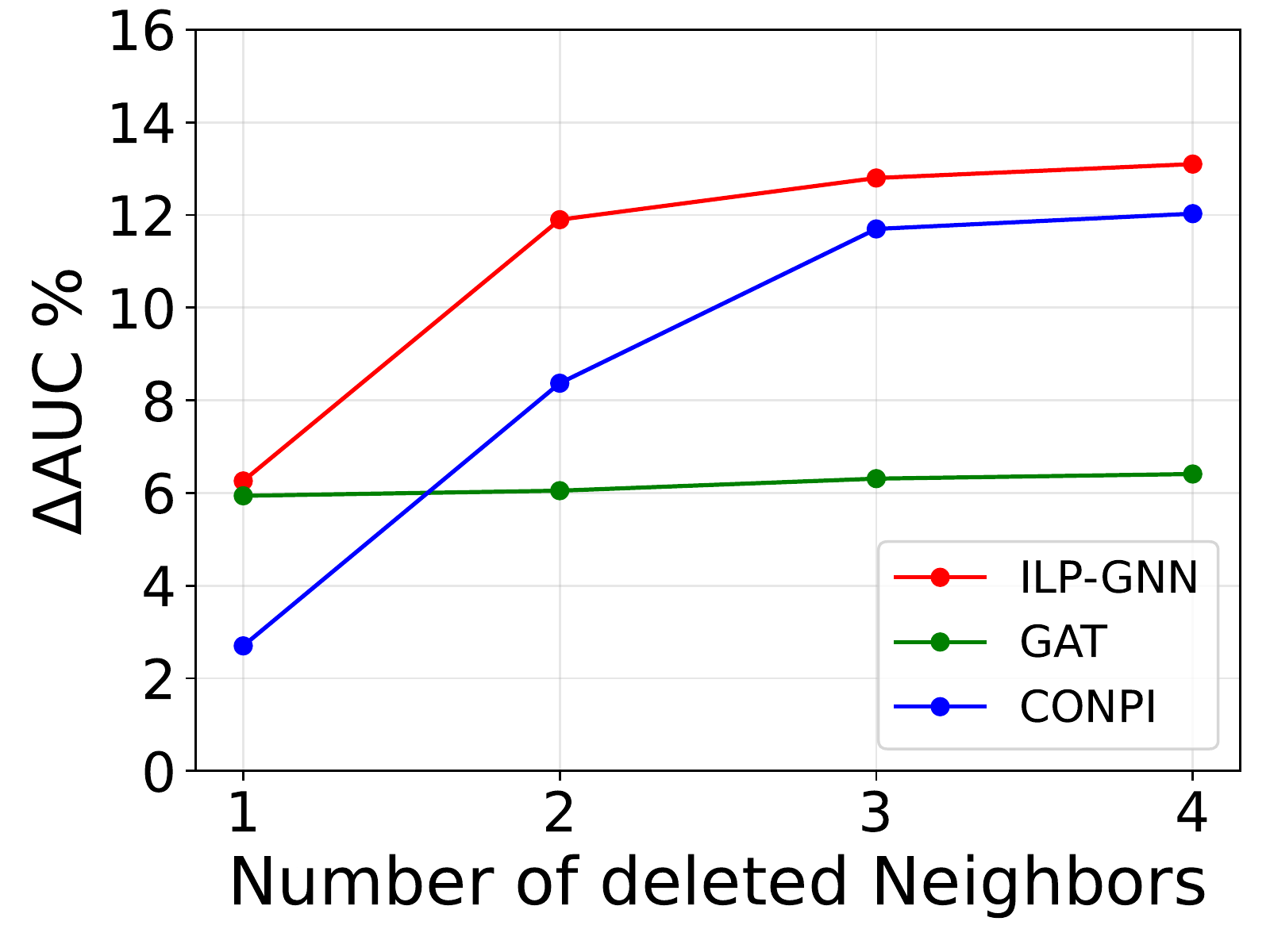}
%\caption{fig2}
\end{minipage}%
}%
\centering
\vskip -2em
\caption{Results on fidelity scores.}
\label{fidelity}
\vskip -1.5em

\end{figure}

% \begin{figure}[t]
% \centering
% \begin{subfigure}[b]{0.4\textwidth}{
% \centering
% \includegraphics[width=\textwidth]{cora_fide.pdf}
% \caption{}
% }%
% \end{subfigure}
% \begin{subfigure}[b]{0.4\textwidth}{
% \centering
% \includegraphics[width=\textwidth]{photo_fide.pdf}
% \caption{}
% }%
% \end{subfigure}
% \centering
% \vskip -2em
% \caption{Results on fidelity scores.}
% \label{fidelity}
% \vskip -1em

% \end{figure}

In this subsection, we conduct quantitatively experimental comparisons and visualization to answer \textbf{RQ2}. 

\subsubsection{Results on Fidelity Scores.} We first demonstrate the effectiveness of explanation in terms of fidelity scores. The fidelity score measures the link prediction performance drop for each pair of nodes when important neighbors of the pair of nodes are removed. Intuitively, if a model can capture important neighbors of a pair of nodes for link prediction, the removal of such neighbors would result in a significant performance drop. Specifically, for each node pair $(v_i, v_j)$ in the test set, take node $v_i$ as an example, we denote the delete top $M$ neighbors based on weight scores given by the model as $\mathcal{N}_i^o$. %We denote these $M$ neighbors $\mathcal{N}_i^o$. 
We delete will  $\mathcal{N}_i^o$ and obtain a new neighbor set $\mathcal{N}_i \backslash \mathcal{N}_i^o$. We then aggregate neighbors $v_c$ of $v_i$ from this new set $v_c \in \mathcal{N}_i \backslash \mathcal{N}_i^o$ to obtain the representation vector $\mathbf{h}_i^{w}$ via different aggregation methods from different models, i.e., GAT, CONPI, ILP-GNN. We do the same operations for  node $v_j$ to get the representation $\mathbf{h}_j^{w}$. 
% \suhang{unclear description. how many nodes?}. \suhang{you need to explain the intuition of fidelity score}
Note that if $|\mathcal{N}_i| \leq M$ and the new neighbors set is empty, we only use their features. Then, we can obtain the new link prediction score $\hat{p}^{w}_{ij}$ using $\mathbf{h}_i^{w}$ and $\mathbf{h}_j^{w}$ by Eq.(\ref{eq:8}). The fidelity score $\Delta AUC\%$ is calculated as $\Delta AUC\% = \sum_{e_{ij}\in \mathcal{E}_U} (AUC_{\ \hat{p}_{ij}} - AUC_{\hat{p}^{w}_{ij}})\%$, where $AUC_{\ \hat{p}_{ij}}$ and $AUC_{\ \hat{p}^{w}_{ij}}$ are AUC value with regard to different predicted results, and $\hat{p}_{ij}$ is the original predicted result. We compare our model with GAT and CONPI which can also assign weights to neighbors for link prediction. We vary the number of deleted neighbor $M$ as $\{1,2,3,4\}$. All experiments are conducted five times with random splits and the results are reported in Figure~\ref{fidelity}. From the figure, we make the following observations: (i) ILP-GNN consistently outperforms two baselines with different deleted neighbors. Compared with GAT, for different links $(v_i, v_j)$ and $(v_i, v_k)$, ILP-GNN learns pair specific representation for $v_i$ by selecting different neighbors of $v_i$ relevant to the existence of various links. However, GAT assigns higher weights to the same neighbors and learns one representation of $v_i$ for different links. In real-word graphs, different links may be from different factors, which are relevant to different neighbors of nodes. Therefore, our proposed ILP-GNN can explore the relevant neighbors for links with different factors, which will greatly improve link prediction and lead to higher fidelity scores. (ii) Also, our proposed loss function can help the model select neighbors to give high probabilities to links and low probabilities to no-links. It can help the model find neighbors relevant to links. Thus, our model can achieve the best fidelity score by deleting neighbors.

\begin{table}[t!]
\centering
\caption{Explanation Performance on Synthetic Dataset}
\vspace{-4.5mm}
\resizebox{1.02\columnwidth}{!}{
	\begin{tabular}{l|cc|cc|cc}
	\toprule
	\multirow{2}*{\textbf{{Method}}} & \multicolumn{2}{c} {\textbf{Syn-sparse}} & \multicolumn{2}{c} {\textbf{Syn-medium}} & \multicolumn{2}{c} {\textbf{Syn-dense}} \\
	\cline{2-3} \cline{4-5} \cline{6-7}
% 	\cline{2-4} \cline{5-7} \cline{8-10}
	& Precision@1 & Precision@2 & Precision@1 & Precision@2 & Precision@1 & Precision@2\\
	\hline
    Random & 30.62 $\pm$ 1.77& 31.96 $\pm$ 1.03 & 26.44$\pm$ 1.05 & 25.85$\pm$ 0.93 & 12.75$\pm$ 0.92 & 13.15 $\pm$ 0.82 \\
	GAT & 33.92 $\pm$ 2.21 & 39.40 $\pm$ 1.68 & 28.39 $\pm$ 1.58 & 28.17 $\pm$ 1.31 & 24.05 $\pm$ 2.34 & 23.74 $\pm$ 1.06 \\
	CONPI & 39.29 $\pm$ 2.53 & 46.22 $\pm$ 2.03 & 30.75 $\pm$ 2.53 & 15.23 $\pm$ 4.12 & 32.46 $\pm$ 1.21 & 19.22 $\pm$ 1.03 \\
     GNNExplainer & 24.46 $\pm$ 1.97 & 34.01 $\pm$ 1.72 & 27.22 $\pm$ 1.97 & 32.75 $\pm$ 1.85 & 29.90 $\pm$ 1.84 & 35.60 $\pm$ 1.89 \\
	ILP-GNN & \textbf{49.68 $\pm$ 3.87} & \textbf{65.25 $\pm$ 3.49} & \textbf{81.27 $\pm$ 2.67} & \textbf{84.31 $\pm$ 1.76} &  \textbf{79.10 $\pm$ 0.15} & \textbf{82.22 $\pm$ 0.30} \\
	\bottomrule
	\end{tabular}
	}
	\label{tab:syn_results}
	\vspace{-3mm}
\end{table}

\begin{table}[t]
    \small
    \centering
    \caption{Link Prediction Performance (AUC \%) on Synthetic Dataset.}
    \vskip -1.5em
    \resizebox{1.02\columnwidth}{!}{
    \begin{tabular}{ccccc}
    \toprule
    \textbf{Method}  & GCN & GAT & CONPI & ILP-GNN\\
    \midrule
    Syn-sparse & 78.88 $\pm$ 1.46 & 79.11 $\pm$ 1.70 & 78.93 $\pm$ 1.51 & \textbf{80.42 $\pm$ 0.58}\\
    % Neigh. ACC & 69.2$\pm 4.0$ & 47.9$\pm 0.9$ & 33.0$\pm1.8$ & \textbf{95.5}$\pm \mathbf{0.2} $\\
    Syn-medium & 82.92 $\pm$ 1.90 & 83.04 $\pm$ 1.86 & 83.18 $\pm$ 1.72 & \textbf{83.44 $\pm$ 1.76} \\
    Syn-dense & 23.74 $\pm$ 1.06 & 84.64 $\pm$ 1.80 & 84.23 $\pm$ 2.67 & \textbf{84.85 $\pm$ 0.42} \\
    \bottomrule
    \end{tabular}
    }
    \label{tab:syn_link}
    \vskip -1em
\end{table}

\subsubsection{Results on Synthetic Datasets} Secondly, we evaluate the explanation quality on synthetic datasets with groundtruth explanation.  Specifically, in the synthetic datasets, for each node pair $(v_i, v_j)$, $v_i$ is similar to $K$ neighbors of node $v_j$ and $v_j$ are similar to $K$ neighbors of $v_i$, which lead to the link between them. We treat these $K$ neighbors of $v_i$ and $K$ neighbors of $v_j$ as explanation neighbors for the link between them. Therefore, for the task to find explanation neighbors in the synthetic datasets, the model should find the correct $K$ neighbors of $v_i$ and $K$ neighbors of $v_j$ for explainable link prediction.  Specifically, for each node pair $(v_i, v_j)$, we rank neighbors of $v_i$ based on weight scores assigned from the model, i.e., ILP-GNN, GAT and CONPI.  %Then, we treat list of explanation neighbors of $v_i$ in the synthetic datasets as groundtruth. 
We also do the same operations on $v_j$. We then calculate the precision@2 and precision@1 for the ranked neighbors list of $v_i$ and $v_j$ w.r.t groundtruth explanation neighbors set of $v_i$ and $v_j$. Also, we includes the baseline (named Random in Table~\ref{tab:syn_results}) which randomly select $K$ neighbors of $v_i$ and $v_j$ and evaluate the explanation performance based on randomly selected $K$ neighbors list. To further demonstrate the effectiveness of our model on link prediction explanation, we adopt one classical post-hoc explanation method, GNNExplainer, which is originally designed for node classification. For the link between $v_i$ and $v_j$, we adopt GNNExplainer to find the top $K$ crucial neighbors of them for the link between them.  We treat these $K$ neighbors of $v_i$ and $v_j$ as explanation neighbors of them.
The higher the precision@1 or precision@2 is, the closer the explainable neighbors found by the model with the groundtruth. We evaluate explanation performance by precision@1 or precision@2 and link prediction performance by AUC. All experiments are conducted five times and the average results and standard deviations on the three synthetic datasets are reported in Table~\ref{tab:syn_results} and Table~\ref{tab:syn_link} for explanation and link prediction performance. From the results, we make the following observations: (i) GNNExplainer only has a little improvement compared with the Random method. It demonstrates that current explanation methods designed for node classification can't be easily adopted to link prediction. (ii) ILP-GNN consistently outperforms all baselines on both explanation and link prediction metrics, which indicates that it can retrieve reliable $K$ neighbor nodes for prediction and explanation simultaneously.

\begin{figure*}[t]
\centering
\subfigure[A link between nodes $v_1$ and $v_2$ without common neighbors]{
\begin{minipage}[t]{0.2\linewidth}
\centering
\vskip -2pt 
\includegraphics[width=\textwidth]{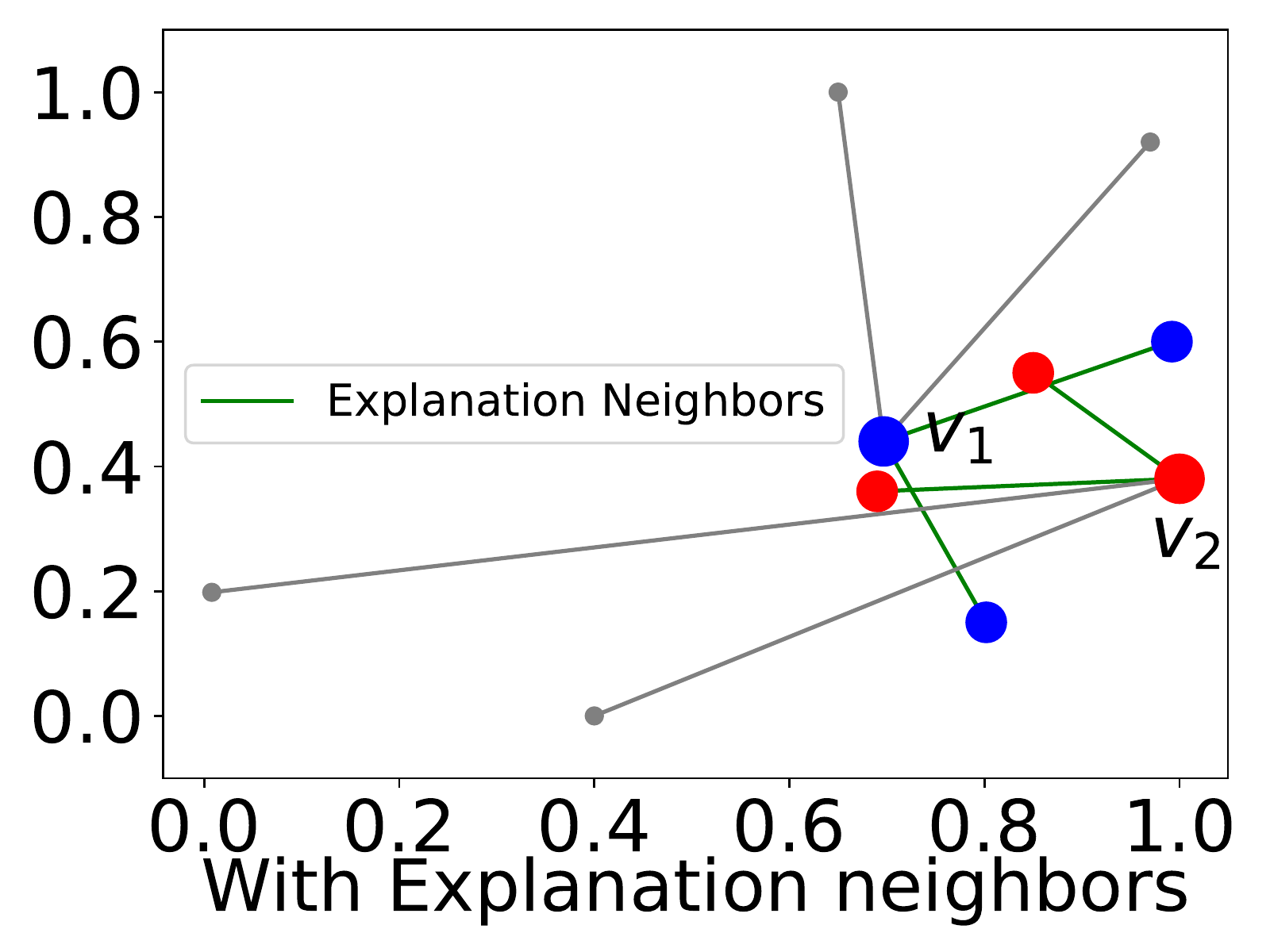}
%\caption{fig1}
\end{minipage}%
\begin{minipage}[t]{0.2\linewidth}
\centering
\vskip -2pt 
\includegraphics[width=\textwidth]{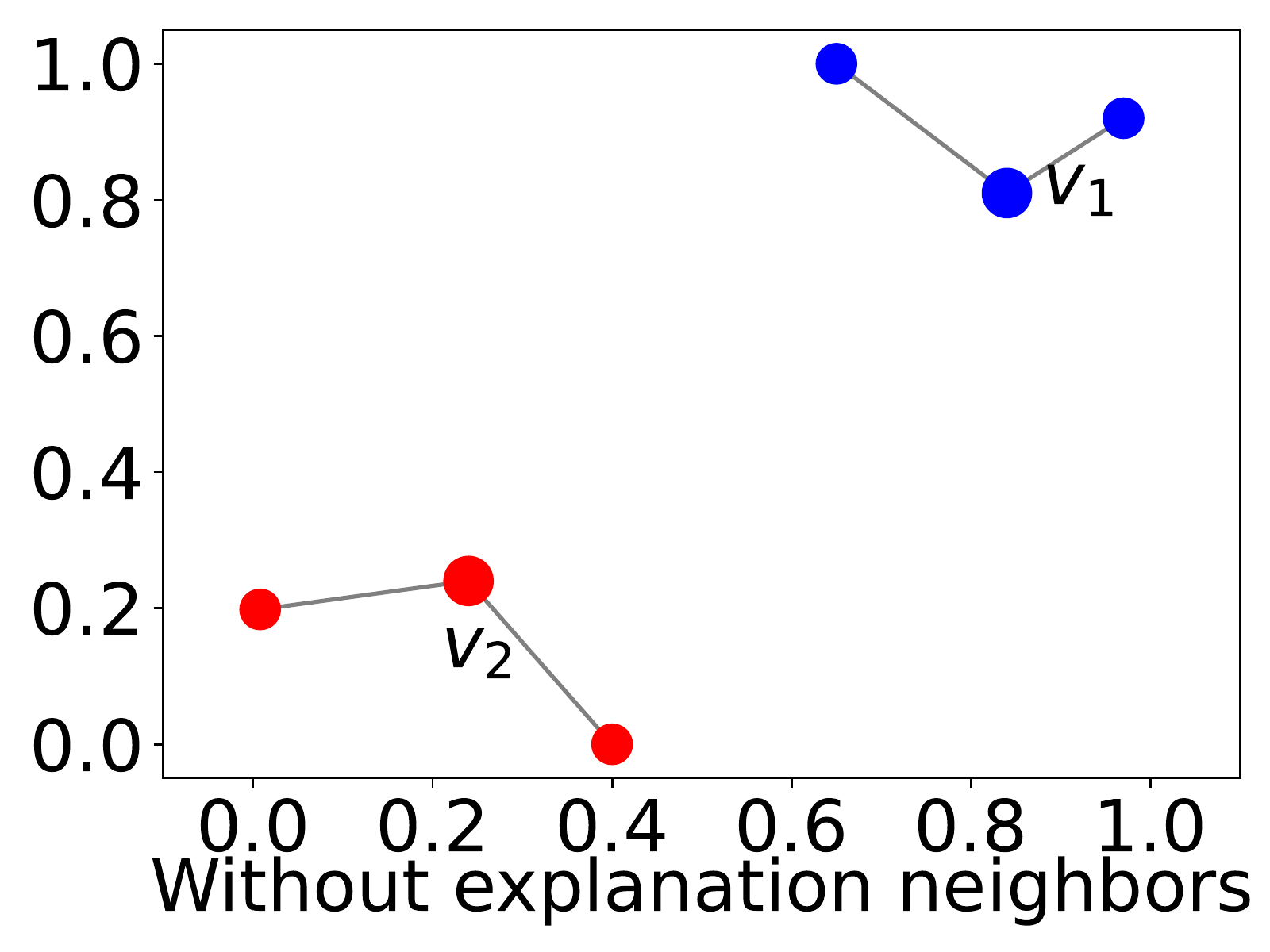}
%\caption{fig2}
\end{minipage}%
}%
\subfigure[A link between nodes $v_1$ and $v_2$ with common neighbors]{
\begin{minipage}[t]{0.2\linewidth}
\centering
\vskip -2pt 
\includegraphics[width=\textwidth]{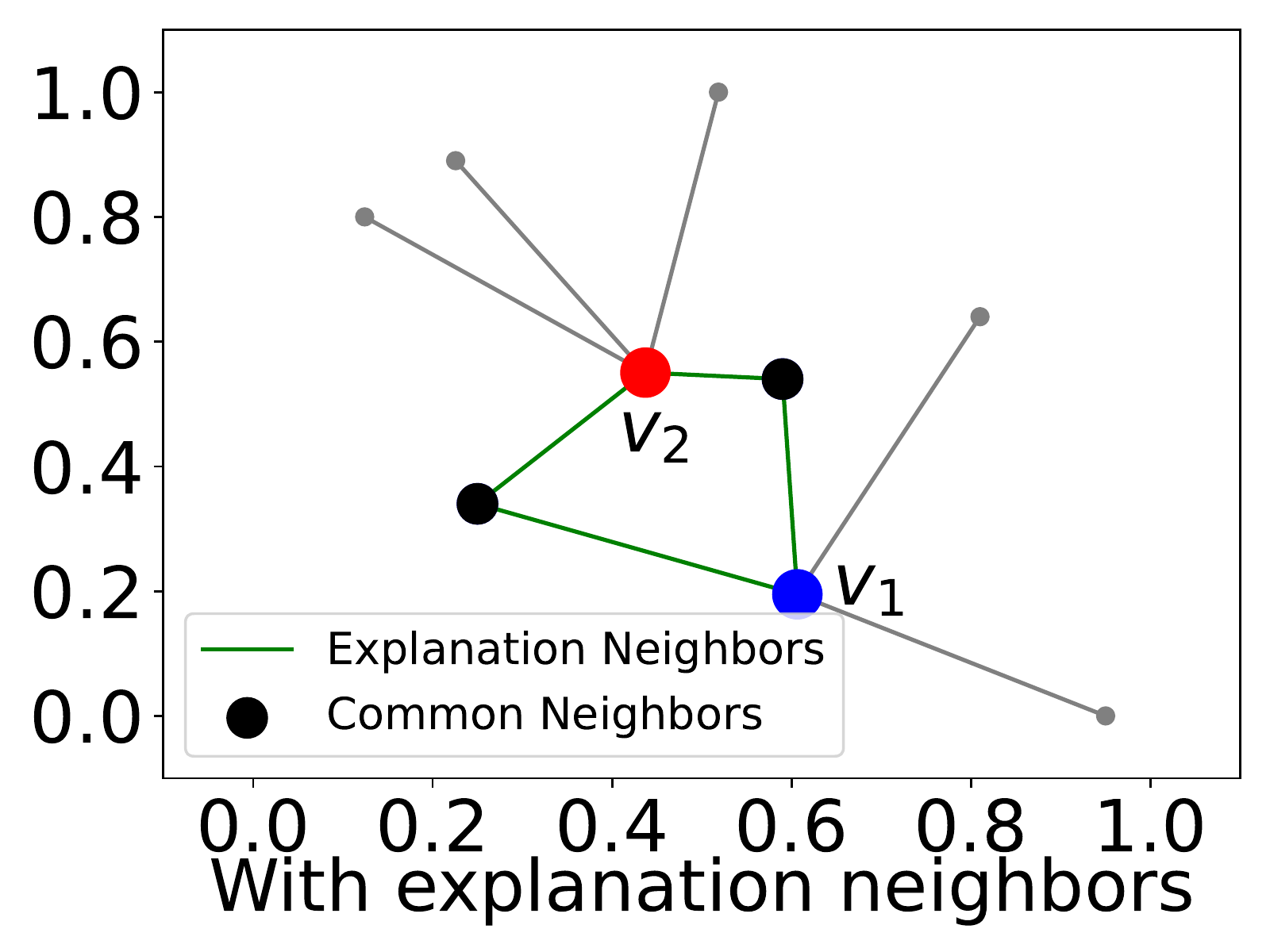}
%\caption{fig1}
\end{minipage}%
\begin{minipage}[t]{0.2\linewidth}
\centering
\vskip -2pt 
\includegraphics[width=\textwidth]{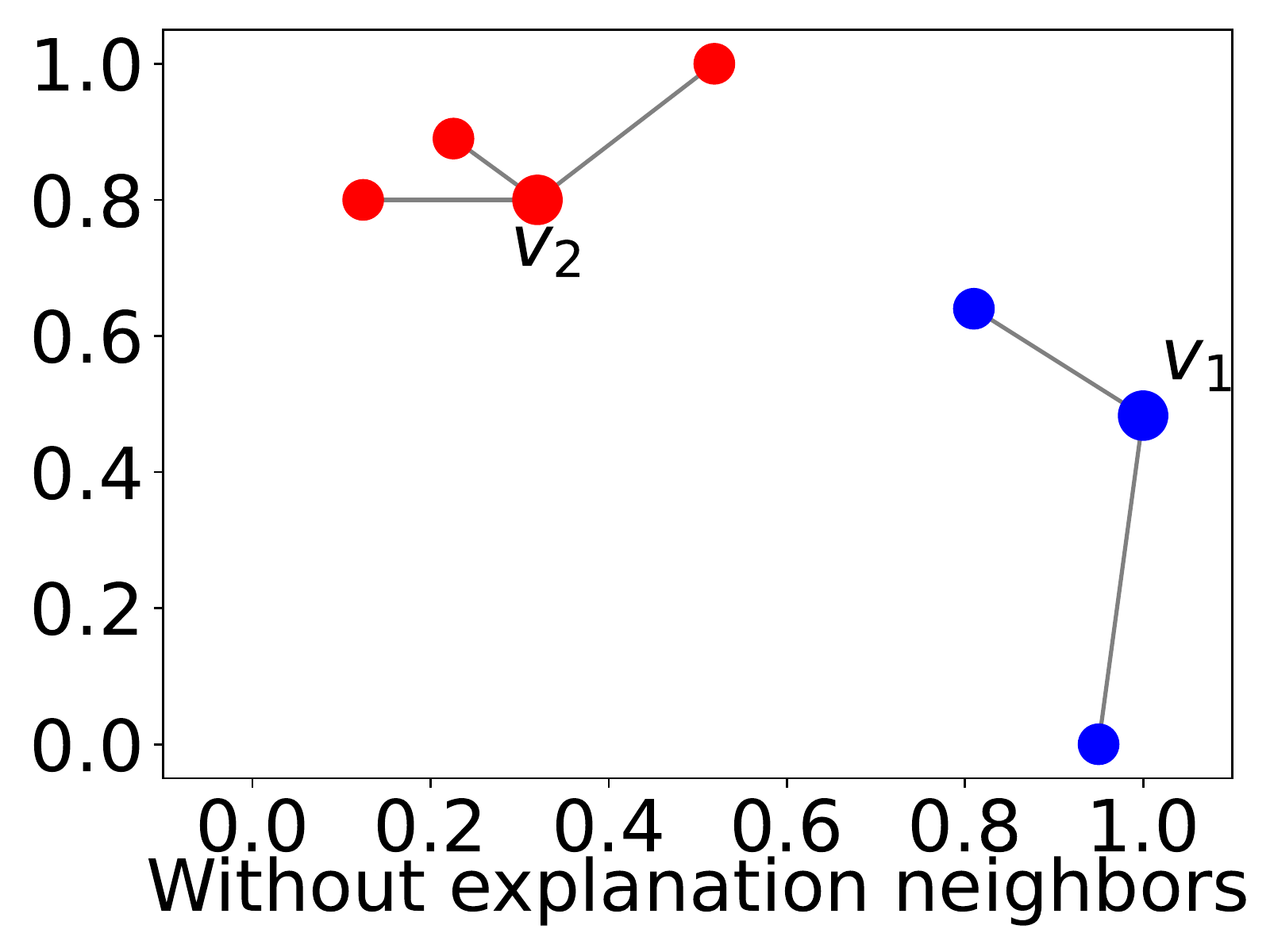}
%\caption{fig2}
\end{minipage}%
}%
\subfigure[No link between nodes $v_1$ and $v_2$]{
\begin{minipage}[t]{0.2\linewidth}
\centering
\vskip -2pt 
\includegraphics[width=\textwidth]{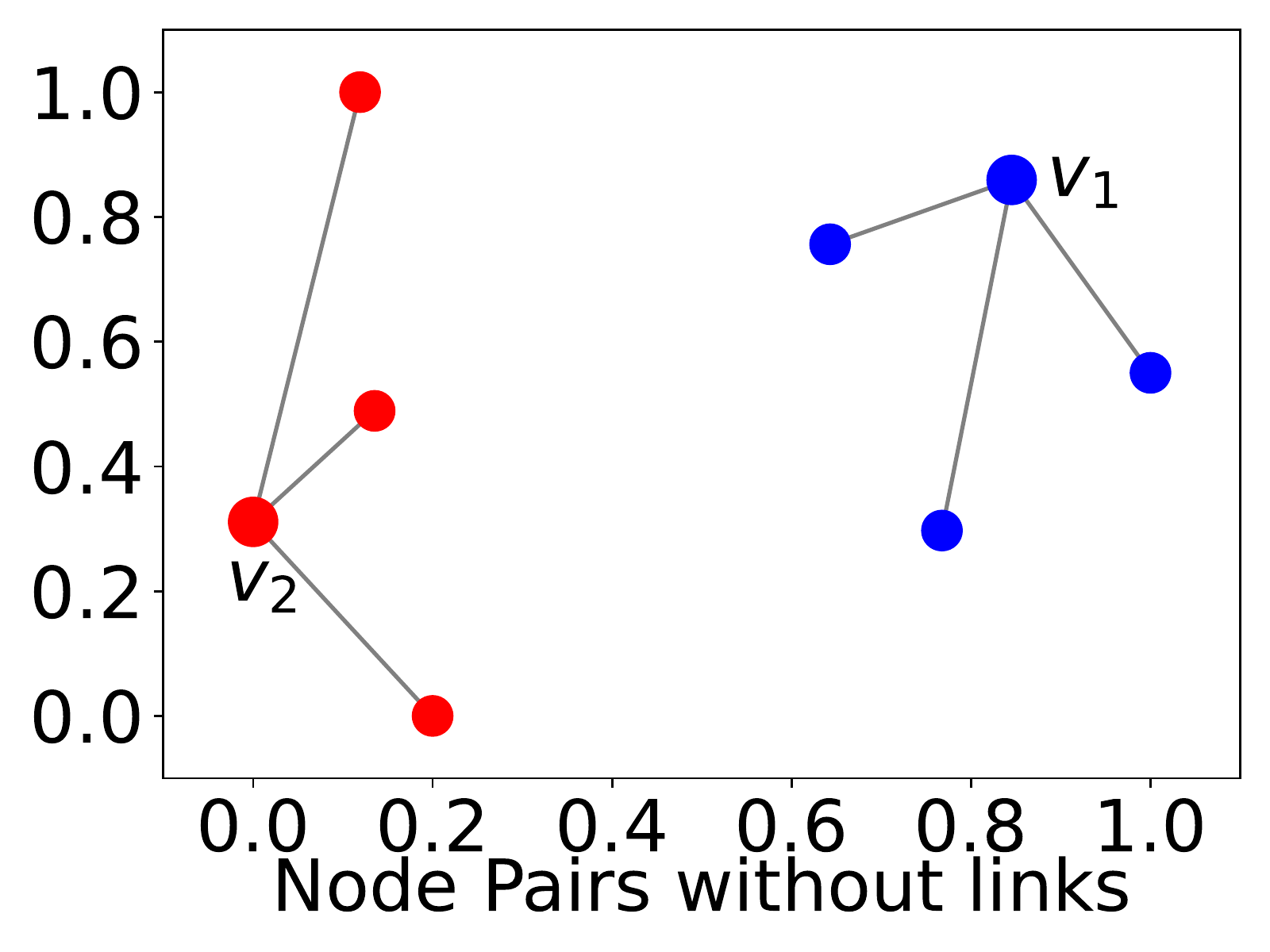}
%\caption{fig1}
\end{minipage}%
}%
\centering
\vskip -2em
\caption{Case Study for node pairs with links or without links on Cora.}
\label{case_study}
\vskip -1em
\end{figure*} 

\begin{figure}[t]
\centering
\subfigure[Cora]{
\begin{minipage}[t]{0.42\linewidth}
\centering
\vskip -2pt 
\includegraphics[width=\textwidth]{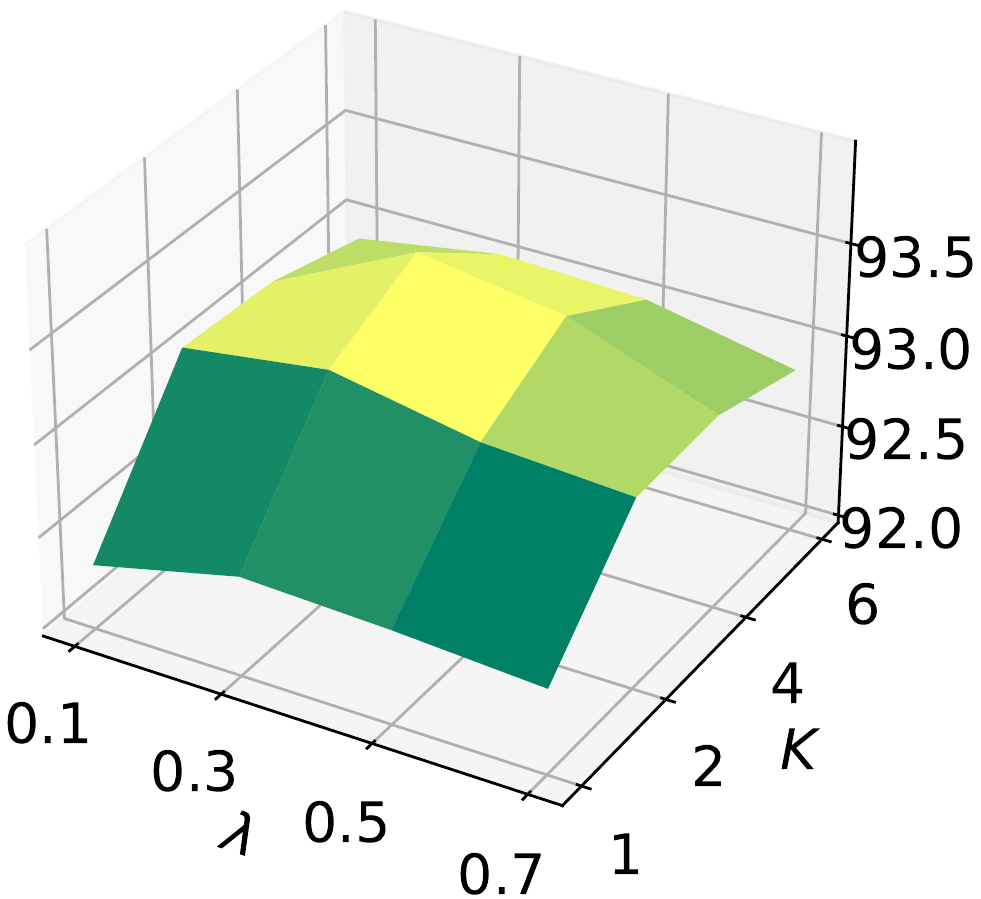}
%\caption{fig1}
\end{minipage}%
}~~~~\subfigure[Citeseer]{
\begin{minipage}[t]{0.42\linewidth}
\centering
\vskip -2pt 
\includegraphics[width=\textwidth]{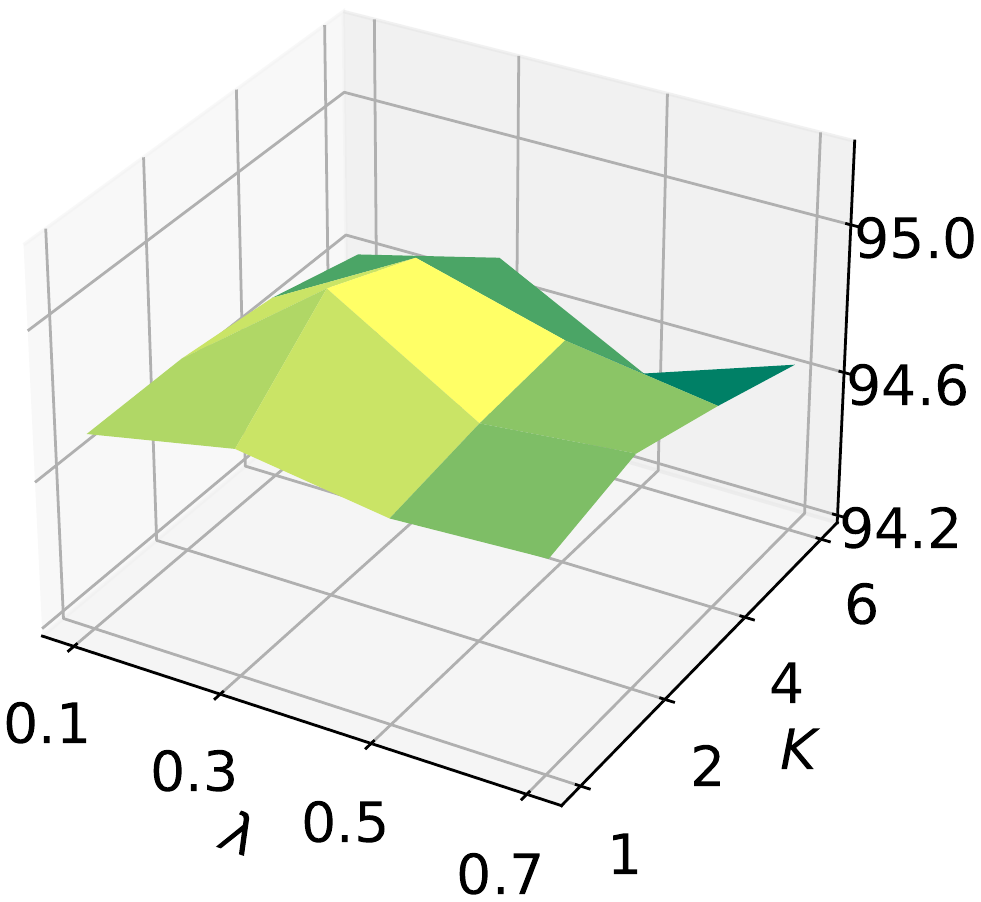}
%\caption{fig2}
\end{minipage}%
}%
\centering
\vskip -2em
\caption{Hyperparameter Sensitivity Analysis}
\label{hyper}
\vskip -1em
\end{figure}

\begin{figure}[t]
\centering
\subfigure[ILP-GNN]{
\begin{minipage}[t]{0.46\linewidth}
\vskip -2.2pt 
\centering
\includegraphics[width=\textwidth]{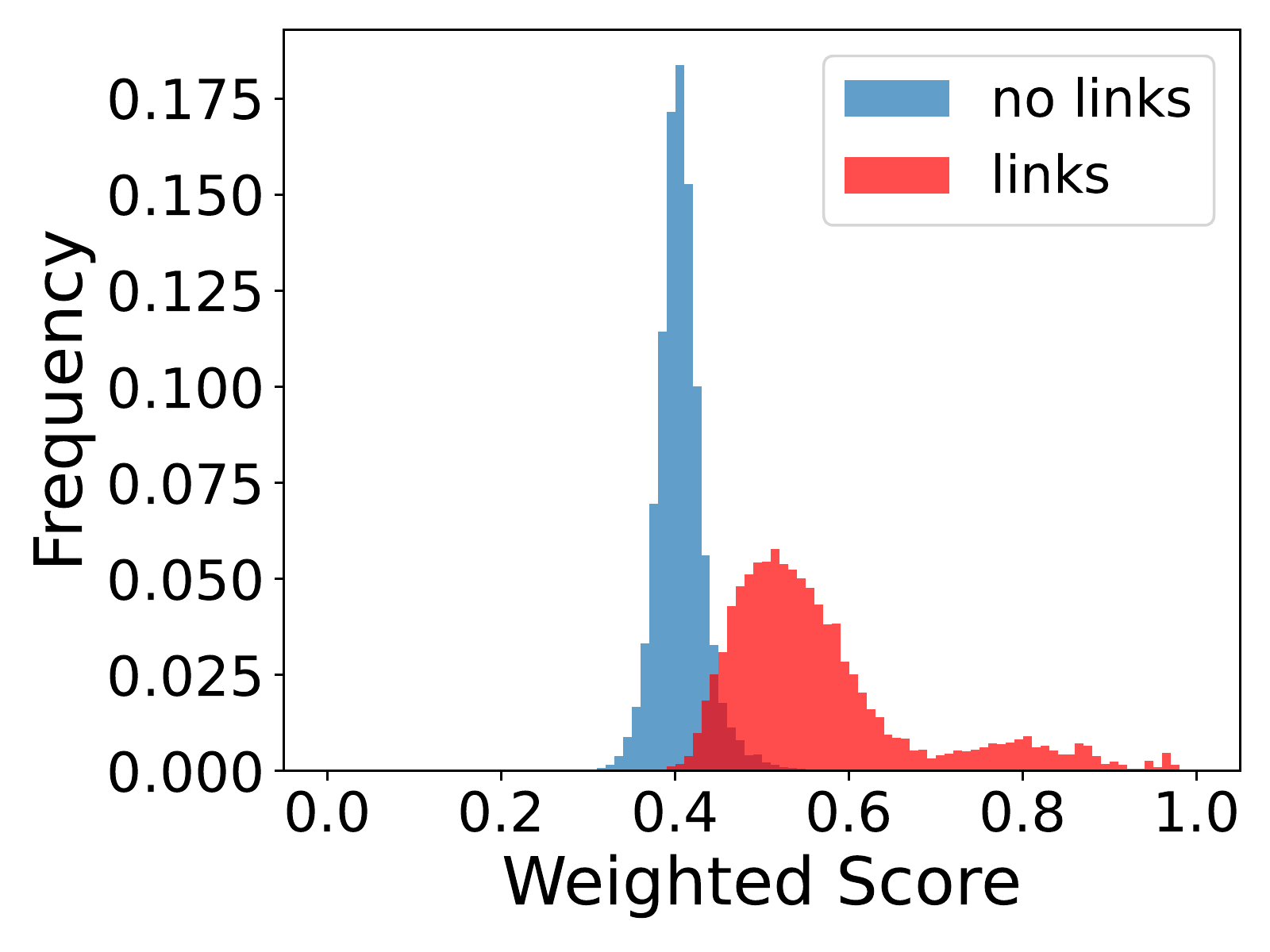}
%\caption{fig1}
\end{minipage}%
}~~%
\subfigure[CONPI]{
\begin{minipage}[t]{0.46\linewidth}
\vskip -2.2pt 
\centering
\includegraphics[width=\textwidth]{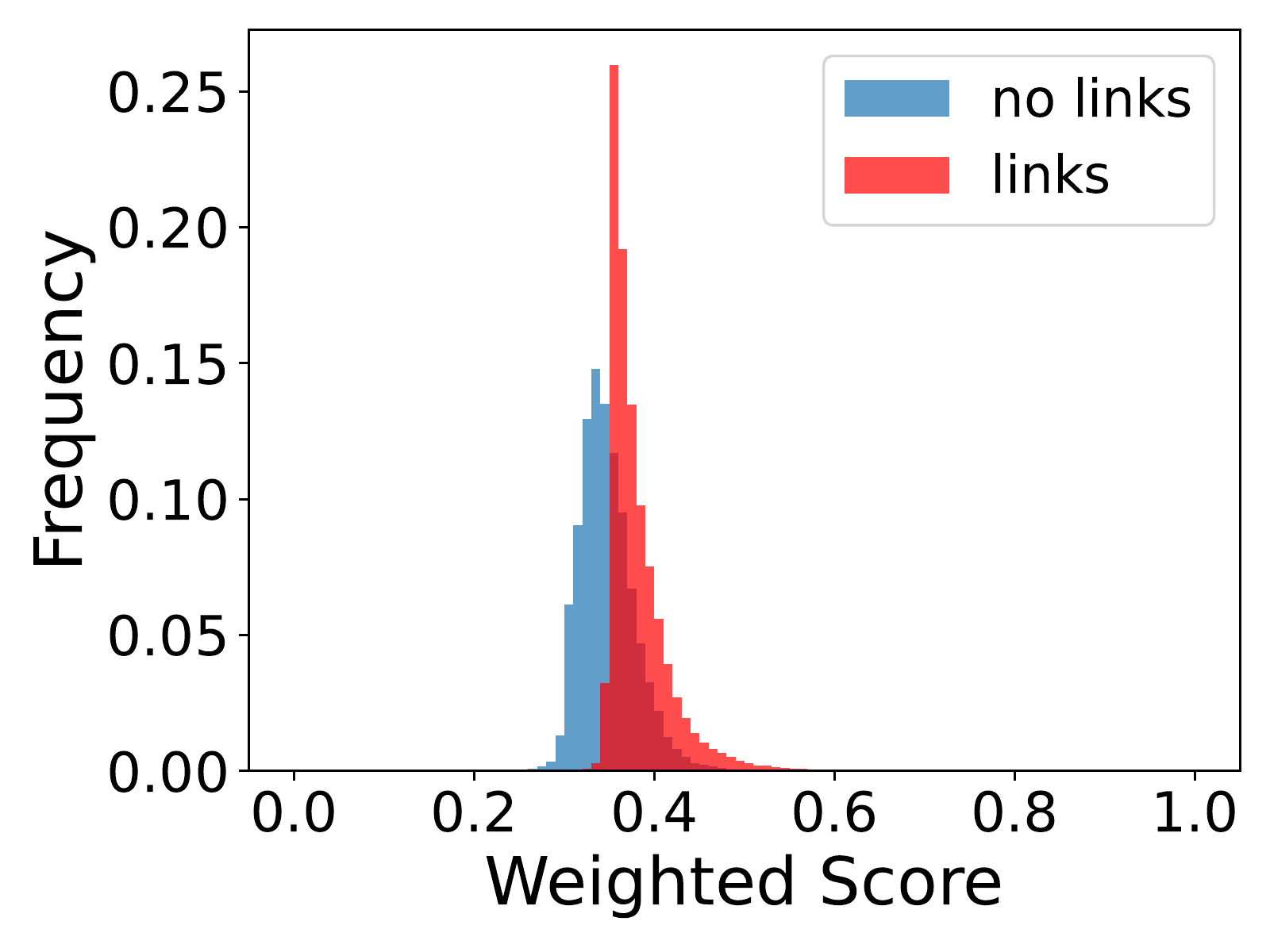}
%\caption{fig2}
\end{minipage}%
}%
\centering
\vskip -2em
\caption{Visualization of weighted scores.}
\label{weighted_scores}
\vskip -1em

\end{figure}
\subsubsection{Visualization of Weights.} Finally, we visualize the distribution of learned weight scores for our model and CONPI on both existent links and no-links. The experiment is conducted on Cora and the results are shown in Figure~\ref{weighted_scores}. We can observe that  CONPI can't recognize different neighbors for links and no-links, which shows the reason for their bad performance in Table~\ref{main_results}. For each linked node pair $(v_i, v_j)$, our model assigns higher weights to neighbors of $v_i$ that indicate neighbors of it are similar to $v_j$. It will result in higher predicted probabilities by aggregating these neighbors to learn representation $\mathbf{h}_i$ and then predict probabilities for this link in Eq.(\ref{eq:8}). 
Also, for dissimilar neighbors, ILP-GNN will give lower probabilities to no-links. Therefore, it further demonstrates the effectiveness of our model to select relevant neighbors to improve the performance of link prediction.

 \begin{figure}[t]
\centering
% \captionsetup[subfigure]{aboveskip=-2pt,belowskip=-2pt}
\subfigure[Cora]{
\begin{minipage}[t]{0.44\linewidth}
\vskip -2pt 
\centering
\includegraphics[width=\textwidth]{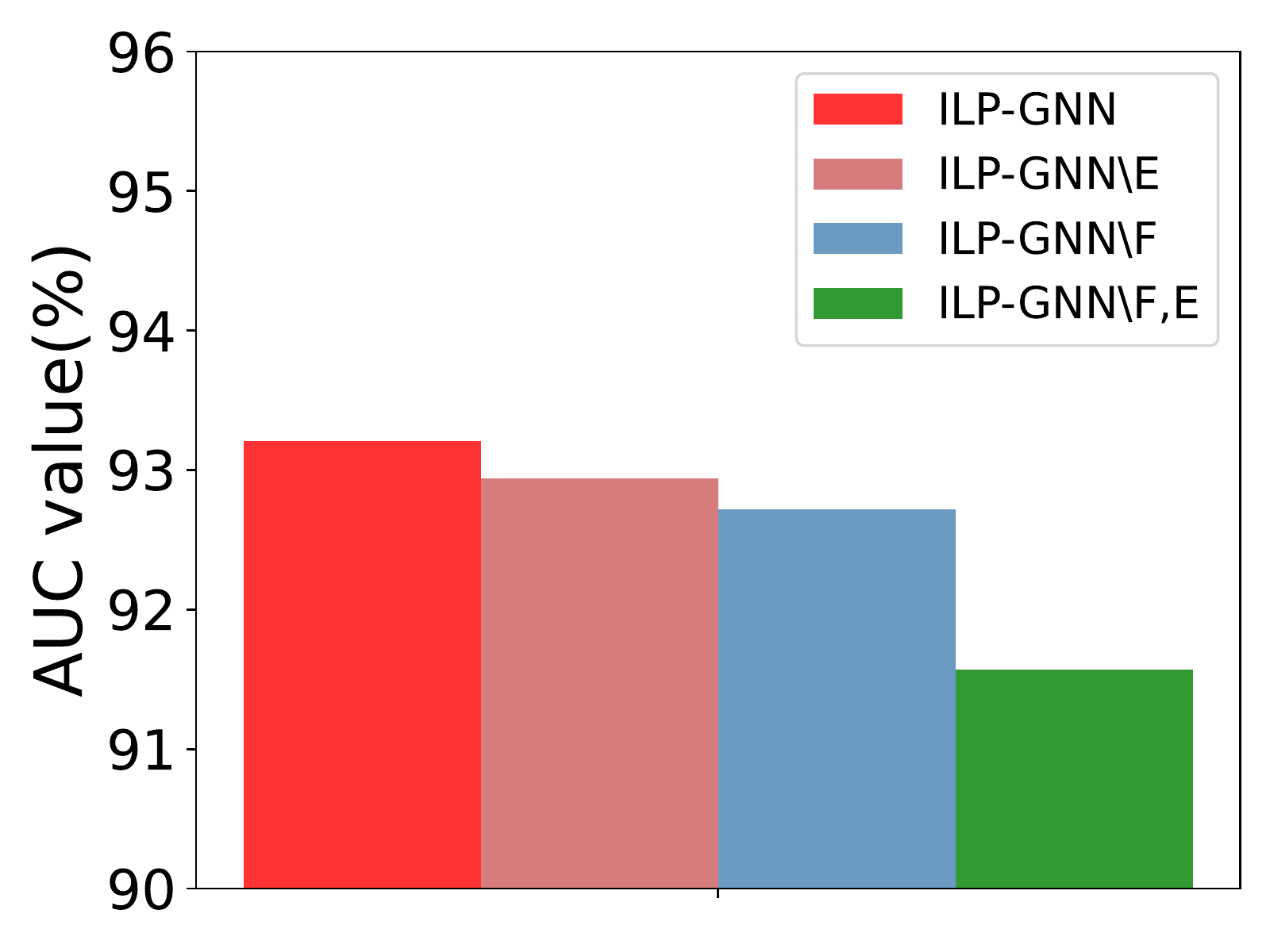}
%\caption{fig1}
\end{minipage}%
}~~~~%\hspace{-20pt}
\subfigure[Photo]{
\begin{minipage}[t]{0.44\linewidth}
\vskip -2pt 
\centering
\includegraphics[width=\textwidth]{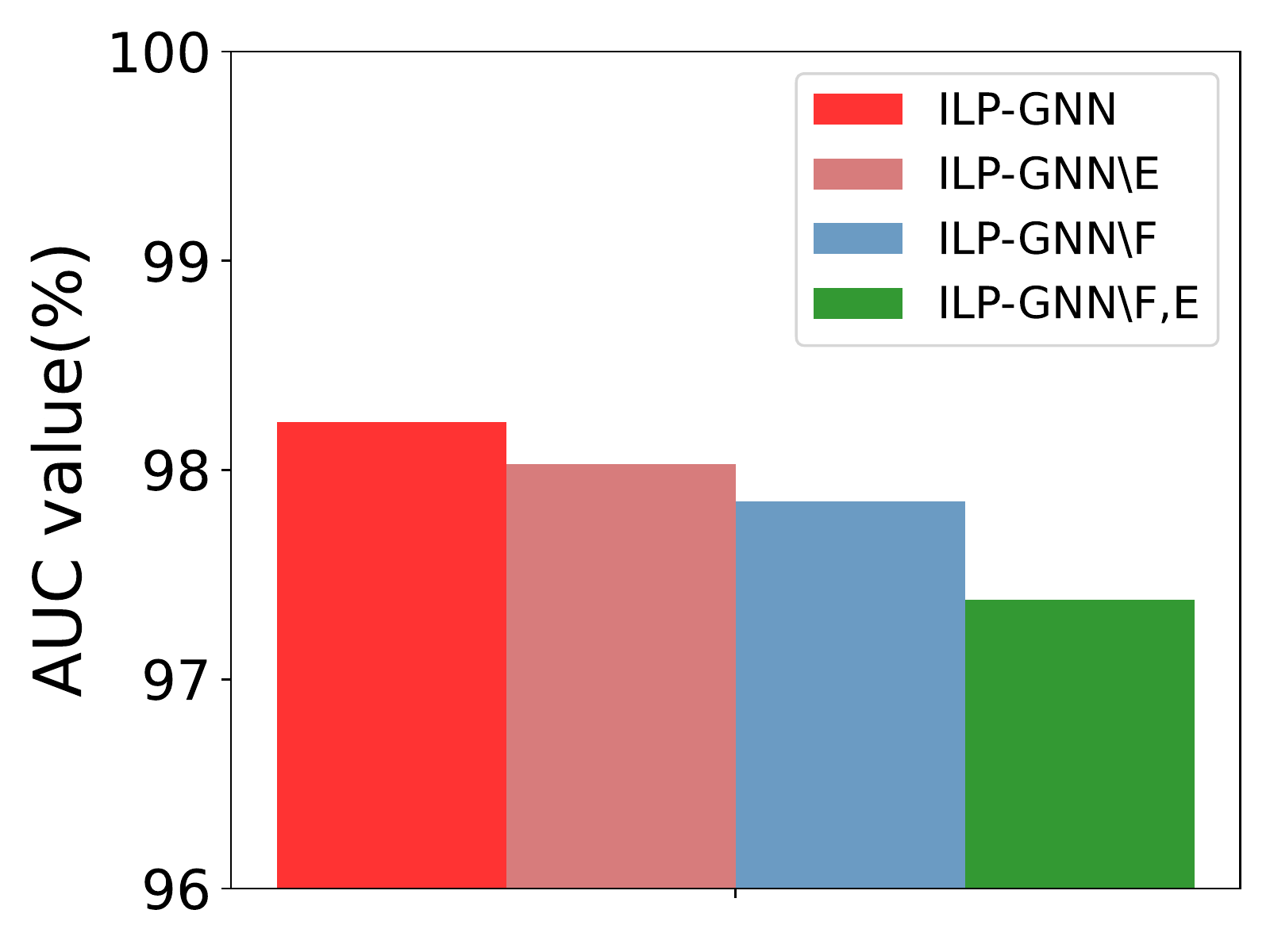}
%\caption{fig2}
\end{minipage}%
}%
\centering
\vskip -2em
\caption{Ablation Study on Cora and Photo datasets.}
\label{ablation_study}
\vskip -2em

\end{figure}

\subsection{Ablation Study}

To answer \textbf{RQ3}, we conduct an ablation study to explore the effects of both node and structure similarity to find $K$ interpretable neighbors for the link prediction task. ILP-GNN$\backslash$E denotes the variant that we don't use structure similarity measurement to find neighbors, i.e., setting $\alpha=0$ in Eq.(\ref{eq:5}). ILP-GNN$\backslash$F denotes the variant without node similarity measurement in Eq.(\ref{eq:5}), i.e., setting $\alpha=1$. ILP-GNN$\backslash$F, E represents the GCN model that aggregates all neighbors without weights. The experimental results on Cora and Photo are reported in Figure~\ref{ablation_study}. We can observe that: (a) two variants of ILP-GNN and ILP-GNN by predicting links with selecting neighbors can greatly outperform ILP-GNN$\backslash$F, E. It means that the proposed interpretable neighbor aggregation module can greatly improve the performance of link prediction by finding $K$ neighbors of one node, which are similar to other linked nodes. (b) ILP-GNN can outperform ILP-GNN$\backslash$F and ILP-GNN$\backslash$E, which implies that combining both node and structure similarity is helpful for finding relevant neighbors to improve the link prediction.

\subsection{Hyperparameter Sensitivity Analysis}

In this section, we explore the sensitivity of the most important hyperparameters $K$ and $\lambda$, which control the number of neighbors to select and the contribution of the Explanation Enhancement module, respectively. Specifically, we vary $K$ as $\{1, 2, 4, 6\}$ and $\lambda$ as $\{0.1, 0.3, 0.5, 0.7\}$. The other settings are the same as the experiment for Table~\ref{main_results}. We report the AUC value for link prediction on Cora and Citeseer. The experiment is conducted five times with random splits and the average results are shown in Figure~\ref{hyper}. From the figure, we observe that: (i) with the increase of $K$, the performance firstly increases and then decreases. When $K$ is small, a small number of relevant neighbors are selected. A small set of neighbors may not reflect the characteristics of nodes to build links with other nodes, which results in poor performance. When $K$ is large, the large number of selected neighbors may contain bias and will cover important characteristics of nodes building links. Therefore, when $K$ is in the range of 2 to 4, the performance is generally good. (ii) with the increasing of $\lambda$, the performance of ILP-GNN tends to firstly increase and then decrease. When $\lambda$ is small, little supervision is received to select neighbors which may result in a high predicted probability for positive samples and low probabilities for negative samples. Also, large $\lambda$ will be dominated by enhancing selected neighbors for link prediction, which can also lead to poor performance. For $\lambda$, a value between 0.3 to 0.5 generally gives a good performance.

\subsection{Case Study}

We conduct a case study to show the importance of selected neighbors for the decision process of link prediction. Specifically, we apply t-SNE to the learned representation of nodes, i.e. $\mathbf{h}_i$ for node $v_i$ in Eq.(\ref{eq:7}), with aggregating selected neighbors from our model. Then, we learn another representation $\mathbf{h}^s_i$ by aggregating neighbors without these selected neighbors. The relative positions of nodes for these two representations are visualized.  Also, we visualize their local 1-hop neighbors to obtain the positions of nodes based on node similarity as shown in Figure~\ref{case_study}. We show cases of links between two nodes with and without common neighbors and node pairs without links. As shown in Figure~\ref{case_study} (a), the learned representation with selected neighbors from our model will have higher similarity in the left figure ($v_1$ is near to $v_2$). However, the learned embedding without these neighbors will result in a lower similarity in the right figure (two nodes are distant). For no-links, our model will make the learned representation of both $v_1$ and $v_2$ with their neighbors distant. And the model will give lower probabilities for no-links. Therefore, our model can provide both predictions and explanations 
accurately by finding significant neighbors.

%% file: appendix.tex
\section{Datasets} \label{app:data}
\begin{table}[t!]\small
\centering
\caption{Statistics of Datasets.
} \label{tab:datasets}
\vskip -1em
\resizebox{0.75\columnwidth}{!}{
\begin{tabular}{c|cccc}
\bottomrule \textbf{Dataset} & \textbf{Nodes} & \textbf{Edges} & \textbf{Features}  \\ \hline
\hline
Cora  & 2708 & 5278 & 1433 \\
Citeseer  & 3327 &  4552 & 3703 \\
Photo  & 7487 &  119043 & 745 \\
Ogbn-arxiv  & 169343 &  1166243 & 128 \\
Ogbl-collab  & 235868 &  1285465 & 128 \\
Syn-sparse  & 1000 & 4243  & 128 \\
Syn-medium  & 1000 & 9576 & 128 \\
Syn-dense  & 1000 & 19826 & 128 \\
\specialrule{.08em}{0pt}{0pt}
\end{tabular}}
\end{table}
We put the dataset statistics into this section, the table includes the number of features, the number of edges,  and the number of nodes for each dataset. For synthetic datasets, we generate three datasets with the same number of nodes but different numbers of edges. And they are shown in  Table~\ref{tab:datasets}. Also, a detailed description of the synthetic datasets is shown below:

\textbf{Synthetic Datasets:} For synthetic data, we assume that feature of each node is sampled from a Guassian Mxiture Model (GMM) $p(x_i) = \sum_{j=1}^M \phi^i_j \mathcal{N}(\mu_j, I)$ with $M$ components and different weights for each component. Specifically, $N$ nodes are divided into $M$ groups which represent components of GMM  with equal number of nodes, and assign each node with labels $y_i \in \mathbb{R}^M$ which is a one-hot vector and represents which group it belongs to. Then, we add random noise $\delta_i \in \mathbb{R}^M$, where the value of each dimension is sampled from a uniform distribution $U(0, 0.1)$ and obtain the weight vector $\phi^i_j = \text{normalize}(y_i + \delta)$. Then, we obtain node features $X$ sampled from the above Guassian Mixture Model. Based on $X$, we generate the edge between node $v_i$ and $v_j$ via $e_{ij} \sim \text{Bern}((\text{cos}(x_i, x_j)+1)/2)$ with Bernoulli distribution to otain graph $\mathcal{G}_1$, which preserves the homophilious properties of graphs. Then, we expand $\mathcal{G}_1$ to $\mathcal{G}_2$ with explainibility for the existence of edges. Note that the explanation in this paper is about selecting neighbors which are relevant to the existence of links so we will select $K$ neighbors of each node as groundtruth explanation. Firstly, we assume that if a node $v_i$ is similar to neighbors of another connected or unconnected node $v_j$, these two nodes are likely connected. Thus, we can get $s_{\text{NO}}(v_i, v_c, v_j) = \frac{1}{\left|N_i\right|} \sum_{c \in N_i} (\cos (v_c, v_j) + 1)/2$. Secondly, we assume that a node $v_i$ is near to neighbors of another node $v_j$. Specifically, we firstly count the number of paths which are less than 3 via $S_{\text{ST}} = A+\frac{1}{2} A^2+\frac{1}{3} A^3$ and normalize it by $S_{\text{ST}} = \frac{S_{\text{ST}}}{\text{max}(S_{\text{ST}})}$. Then, $s_{\text{ST}}(v_i, v_c, v_j) = \frac{1}{\left|N_i\right|} \sum_{c \in N_i} S_{\text{ST}}^{c, j}$. Finally, top $K$ neighbors based on $(1-\alpha) S_{\text{ST}}^{c, j} + \alpha \cos (v_c, v_j)$ are selected as explanation.  are selected as explanation for the existence of this edge Due to undirected properties of the graph in this paper, we further calculate probability for the existence of edges based on two assumptions $\bar{s}_{\text{NO}}^{i, j} = (s_{\text{NO}}(v_i, v_c, v_j) + s_{\text{NO}}(v_j, v_c, v_i))/2$ and $\bar{s}_{\text{ST}}^{i, j} = (s_{\text{ST}}(v_i, v_c, v_j) + s_{ST}(v_j, v_c, v_i))/2$. Finally, we sample edges via $e_{ij} \sim \text{Bern}(\alpha * \bar{s}_{\text{ST}}^{i, j} + (1- \alpha) * \bar{s}_{\text{ST}}^{i, j})$ to obtain expanded graph $\mathcal{G}_2$. We generate three synthetic datasests with different number of edges for $\alpha =0.3$ and $K$ set as $2, 3, 4$ separately. we randomly select $40\%$ edges from $\mathcal{G}_2 - \mathcal{G}_1$ as testing set and $10\%$ as validation set.

\section{Running Time Comparison} \label{app:time}

We also conduct experiments to compare our training time with baselines. For comparison, we consider three representative and state-of-the-art baselines that achieve great performance on link prediction, including CONPI, SEAL and WP. For SEAL and WP, they sample one subgraph for each link and we also include this sampling process in the running time calculation.  We conduct all running time experiments on the same 64-bit machine with Nvidia GPU (NVIDIA RTX A6000, 1410MHz , 48 GB memory). Furthermore, we count running time of the models with multiple epochs which achieve the best results.  Table~\ref{tab:time}  shows the results of running time for all models. We can observe that our model (ILP-GNN) has a shorter running time than CONPI and SEAL on most of datasets. Our model has a shorter running time on larger datasets than WP, like photo. Also, we have comparable running time with regard to WP on Cora and Citeseer. 

\begin{table}[!t]
    \centering
    \caption{Comparison of the running time where "min" indicates minutes.}
    \label{tab:time}
    {
    \begin{tabular}{lccccccc}
        \toprule[1pt]
    \textbf{Dataset} & \textbf{CONPI} &  \textbf{SEAL} &   \textbf{WP}&  \textbf{ILP-GNN}   \\
    \midrule
         Cora & 4.55 min & 6.33 min &  4.47 min   & 5.10 min  \\
         Citeseer & 5.14 min & 5.64 min  & 4.53 min    & 4.71 min  \\
         Photo & 51.4 min & 48.15 min &  60.33 min   &  48.6 min \\
   \bottomrule[1pt]
    \end{tabular}
    }
\end{table}

%% file: sample-base.bbl

\begin{thebibliography}{45}

%%% the Web address.

\ifx \showCODEN    \undefined \def \showCODEN     #1{\unskip}     \fi
\ifx \showDOI      \undefined \def \showDOI       #1{#1}\fi
\ifx \showISBNx    \undefined \def \showISBNx     #1{\unskip}     \fi
\ifx \showISBNxiii \undefined \def \showISBNxiii  #1{\unskip}     \fi
\ifx \showISSN     \undefined \def \showISSN      #1{\unskip}     \fi
\ifx \showLCCN     \undefined \def \showLCCN      #1{\unskip}     \fi
\ifx \shownote     \undefined \def \shownote      #1{#1}          \fi
\ifx \showarticletitle \undefined \def \showarticletitle #1{#1}   \fi
\ifx \showURL      \undefined \def \showURL       {\relax}        \fi
% invisible to TeX
\providecommand\bibfield[2]{#2}
\providecommand\bibinfo[2]{#2}
\providecommand\natexlab[1]{#1}
\providecommand\showeprint[2][]{arXiv:#2}

\bibitem[Acar et~al\mbox{.}(2009)]%
        {acar2009link}
\bibfield{author}{\bibinfo{person}{Evrim Acar}, \bibinfo{person}{Daniel~M
  Dunlavy}, {and} \bibinfo{person}{Tamara~G Kolda}.}
  \bibinfo{year}{2009}\natexlab{}.
\newblock \showarticletitle{Link prediction on evolving data using matrix and
  tensor factorizations}. In \bibinfo{booktitle}{\emph{2009 IEEE International
  conference on data mining workshops}}. IEEE, \bibinfo{pages}{262--269}.
\newblock


\bibitem[Adamic and Adar(2003)]%
        {adamic2003friends}
\bibfield{author}{\bibinfo{person}{Lada~A Adamic} {and} \bibinfo{person}{Eytan
  Adar}.} \bibinfo{year}{2003}\natexlab{}.
\newblock \showarticletitle{Friends and neighbors on the web}.
\newblock \bibinfo{journal}{\emph{Social networks}} \bibinfo{volume}{25},
  \bibinfo{number}{3} (\bibinfo{year}{2003}), \bibinfo{pages}{211--230}.
\newblock


\bibitem[Brin and Page(1998)]%
        {brin1998anatomy}
\bibfield{author}{\bibinfo{person}{Sergey Brin} {and} \bibinfo{person}{Lawrence
  Page}.} \bibinfo{year}{1998}\natexlab{}.
\newblock \showarticletitle{The anatomy of a large-scale hypertextual web
  search engine}.
\newblock \bibinfo{journal}{\emph{Computer networks and ISDN systems}}
  \bibinfo{volume}{30}, \bibinfo{number}{1-7} (\bibinfo{year}{1998}),
  \bibinfo{pages}{107--117}.
\newblock


\bibitem[Bruna et~al\mbox{.}(2013)]%
        {bruna2013spectral}
\bibfield{author}{\bibinfo{person}{Joan Bruna}, \bibinfo{person}{Wojciech
  Zaremba}, \bibinfo{person}{Arthur Szlam}, {and} \bibinfo{person}{Yann
  LeCun}.} \bibinfo{year}{2013}\natexlab{}.
\newblock \showarticletitle{Spectral networks and locally connected networks on
  graphs}.
\newblock \bibinfo{journal}{\emph{arXiv preprint arXiv:1312.6203}}
  (\bibinfo{year}{2013}).
\newblock


\bibitem[Chen et~al\mbox{.}(2018)]%
        {chen2018fastgcn}
\bibfield{author}{\bibinfo{person}{Jie Chen}, \bibinfo{person}{Tengfei Ma},
  {and} \bibinfo{person}{Cao Xiao}.} \bibinfo{year}{2018}\natexlab{}.
\newblock \showarticletitle{Fastgcn: fast learning with graph convolutional
  networks via importance sampling}.
\newblock \bibinfo{journal}{\emph{arXiv preprint arXiv:1801.10247}}
  (\bibinfo{year}{2018}).
\newblock


\bibitem[Dai and Wang(2021)]%
        {dai2021towards}
\bibfield{author}{\bibinfo{person}{Enyan Dai} {and} \bibinfo{person}{Suhang
  Wang}.} \bibinfo{year}{2021}\natexlab{}.
\newblock \showarticletitle{Towards self-explainable graph neural network}. In
  \bibinfo{booktitle}{\emph{Proceedings of the 30th ACM International
  Conference on Information \& Knowledge Management}}.
  \bibinfo{pages}{302--311}.
\newblock


\bibitem[Fan et~al\mbox{.}(2019)]%
        {fan2019graph}
\bibfield{author}{\bibinfo{person}{Wenqi Fan}, \bibinfo{person}{Yao Ma},
  \bibinfo{person}{Qing Li}, \bibinfo{person}{Yuan He}, \bibinfo{person}{Eric
  Zhao}, \bibinfo{person}{Jiliang Tang}, {and} \bibinfo{person}{Dawei Yin}.}
  \bibinfo{year}{2019}\natexlab{}.
\newblock \showarticletitle{Graph neural networks for social recommendation}.
  In \bibinfo{booktitle}{\emph{The world wide web conference}}.
  \bibinfo{pages}{417--426}.
\newblock


\bibitem[Gao et~al\mbox{.}(2018)]%
        {gao2018large}
\bibfield{author}{\bibinfo{person}{Hongyang Gao}, \bibinfo{person}{Zhengyang
  Wang}, {and} \bibinfo{person}{Shuiwang Ji}.} \bibinfo{year}{2018}\natexlab{}.
\newblock \showarticletitle{Large-scale learnable graph convolutional
  networks}. In \bibinfo{booktitle}{\emph{Proceedings of the 24th ACM SIGKDD
  international conference on knowledge discovery \& data mining}}.
  \bibinfo{pages}{1416--1424}.
\newblock


\bibitem[Glorot and Bengio(2010)]%
        {glorot2010understanding}
\bibfield{author}{\bibinfo{person}{Xavier Glorot} {and} \bibinfo{person}{Yoshua
  Bengio}.} \bibinfo{year}{2010}\natexlab{}.
\newblock \showarticletitle{Understanding the difficulty of training deep
  feedforward neural networks}. In \bibinfo{booktitle}{\emph{Proceedings of the
  thirteenth international conference on artificial intelligence and
  statistics}}. JMLR Workshop and Conference Proceedings,
  \bibinfo{pages}{249--256}.
\newblock


\bibitem[Hamilton et~al\mbox{.}(2017)]%
        {hamilton2017inductive}
\bibfield{author}{\bibinfo{person}{Will Hamilton}, \bibinfo{person}{Zhitao
  Ying}, {and} \bibinfo{person}{Jure Leskovec}.}
  \bibinfo{year}{2017}\natexlab{}.
\newblock \showarticletitle{Inductive representation learning on large graphs}.
\newblock \bibinfo{journal}{\emph{Advances in neural information processing
  systems}}  \bibinfo{volume}{30} (\bibinfo{year}{2017}).
\newblock


\bibitem[He et~al\mbox{.}(2021)]%
        {he2021bernnet}
\bibfield{author}{\bibinfo{person}{Mingguo He}, \bibinfo{person}{Zhewei Wei},
  \bibinfo{person}{Hongteng Xu}, {et~al\mbox{.}}}
  \bibinfo{year}{2021}\natexlab{}.
\newblock \showarticletitle{Bernnet: Learning arbitrary graph spectral filters
  via bernstein approximation}.
\newblock \bibinfo{journal}{\emph{Advances in Neural Information Processing
  Systems}}  \bibinfo{volume}{34} (\bibinfo{year}{2021}),
  \bibinfo{pages}{14239--14251}.
\newblock


\bibitem[Hu et~al\mbox{.}(2020)]%
        {hu2020open}
\bibfield{author}{\bibinfo{person}{Weihua Hu}, \bibinfo{person}{Matthias Fey},
  \bibinfo{person}{Marinka Zitnik}, \bibinfo{person}{Yuxiao Dong},
  \bibinfo{person}{Hongyu Ren}, \bibinfo{person}{Bowen Liu},
  \bibinfo{person}{Michele Catasta}, {and} \bibinfo{person}{Jure Leskovec}.}
  \bibinfo{year}{2020}\natexlab{}.
\newblock \showarticletitle{Open graph benchmark: Datasets for machine learning
  on graphs}.
\newblock \bibinfo{journal}{\emph{Advances in neural information processing
  systems}}  \bibinfo{volume}{33} (\bibinfo{year}{2020}),
  \bibinfo{pages}{22118--22133}.
\newblock


\bibitem[Huang et~al\mbox{.}(2022)]%
        {huang2022graphlime}
\bibfield{author}{\bibinfo{person}{Qiang Huang}, \bibinfo{person}{Makoto
  Yamada}, \bibinfo{person}{Yuan Tian}, \bibinfo{person}{Dinesh Singh}, {and}
  \bibinfo{person}{Yi Chang}.} \bibinfo{year}{2022}\natexlab{}.
\newblock \showarticletitle{Graphlime: Local interpretable model explanations
  for graph neural networks}.
\newblock \bibinfo{journal}{\emph{IEEE Transactions on Knowledge and Data
  Engineering}} (\bibinfo{year}{2022}).
\newblock


\bibitem[Jeh and Widom(2002)]%
        {jeh2002simrank}
\bibfield{author}{\bibinfo{person}{Glen Jeh} {and} \bibinfo{person}{Jennifer
  Widom}.} \bibinfo{year}{2002}\natexlab{}.
\newblock \showarticletitle{Simrank: a measure of structural-context
  similarity}. In \bibinfo{booktitle}{\emph{Proceedings of the eighth ACM
  SIGKDD international conference on Knowledge discovery and data mining}}.
  \bibinfo{pages}{538--543}.
\newblock


\bibitem[Katz(1953)]%
        {katz1953new}
\bibfield{author}{\bibinfo{person}{Leo Katz}.} \bibinfo{year}{1953}\natexlab{}.
\newblock \showarticletitle{A new status index derived from sociometric
  analysis}.
\newblock \bibinfo{journal}{\emph{Psychometrika}} \bibinfo{volume}{18},
  \bibinfo{number}{1} (\bibinfo{year}{1953}), \bibinfo{pages}{39--43}.
\newblock


\bibitem[Kipf and Welling(2016a)]%
        {kipf2016semi}
\bibfield{author}{\bibinfo{person}{Thomas~N Kipf} {and} \bibinfo{person}{Max
  Welling}.} \bibinfo{year}{2016}\natexlab{a}.
\newblock \showarticletitle{Semi-supervised classification with graph
  convolutional networks}.
\newblock \bibinfo{journal}{\emph{arXiv preprint arXiv:1609.02907}}
  (\bibinfo{year}{2016}).
\newblock


\bibitem[Kipf and Welling(2016b)]%
        {kipf2016variational}
\bibfield{author}{\bibinfo{person}{Thomas~N Kipf} {and} \bibinfo{person}{Max
  Welling}.} \bibinfo{year}{2016}\natexlab{b}.
\newblock \showarticletitle{Variational graph auto-encoders}.
\newblock \bibinfo{journal}{\emph{arXiv preprint arXiv:1611.07308}}
  (\bibinfo{year}{2016}).
\newblock


\bibitem[Klicpera et~al\mbox{.}(2019)]%
        {klicpera2019diffusion}
\bibfield{author}{\bibinfo{person}{Johannes Klicpera}, \bibinfo{person}{Stefan
  Wei{\ss}enberger}, {and} \bibinfo{person}{Stephan G{\"u}nnemann}.}
  \bibinfo{year}{2019}\natexlab{}.
\newblock \showarticletitle{Diffusion improves graph learning}.
\newblock \bibinfo{journal}{\emph{arXiv preprint arXiv:1911.05485}}
  (\bibinfo{year}{2019}).
\newblock


\bibitem[L{\"u} et~al\mbox{.}(2009)]%
        {lu2009similarity}
\bibfield{author}{\bibinfo{person}{Linyuan L{\"u}}, \bibinfo{person}{Ci-Hang
  Jin}, {and} \bibinfo{person}{Tao Zhou}.} \bibinfo{year}{2009}\natexlab{}.
\newblock \showarticletitle{Similarity index based on local paths for link
  prediction of complex networks}.
\newblock \bibinfo{journal}{\emph{Physical Review E}} \bibinfo{volume}{80},
  \bibinfo{number}{4} (\bibinfo{year}{2009}), \bibinfo{pages}{046122}.
\newblock


\bibitem[L{\"u} and Zhou(2011)]%
        {lu2011link}
\bibfield{author}{\bibinfo{person}{Linyuan L{\"u}} {and} \bibinfo{person}{Tao
  Zhou}.} \bibinfo{year}{2011}\natexlab{}.
\newblock \showarticletitle{Link prediction in complex networks: A survey}.
\newblock \bibinfo{journal}{\emph{Physica A: statistical mechanics and its
  applications}} \bibinfo{volume}{390}, \bibinfo{number}{6}
  (\bibinfo{year}{2011}), \bibinfo{pages}{1150--1170}.
\newblock


\bibitem[Luo et~al\mbox{.}(2020)]%
        {luo2020parameterized}
\bibfield{author}{\bibinfo{person}{Dongsheng Luo}, \bibinfo{person}{Wei Cheng},
  \bibinfo{person}{Dongkuan Xu}, \bibinfo{person}{Wenchao Yu},
  \bibinfo{person}{Bo Zong}, \bibinfo{person}{Haifeng Chen}, {and}
  \bibinfo{person}{Xiang Zhang}.} \bibinfo{year}{2020}\natexlab{}.
\newblock \showarticletitle{Parameterized explainer for graph neural network}.
\newblock \bibinfo{journal}{\emph{Advances in neural information processing
  systems}}  \bibinfo{volume}{33} (\bibinfo{year}{2020}),
  \bibinfo{pages}{19620--19631}.
\newblock


\bibitem[Maree et~al\mbox{.}(2020)]%
        {maree2020towards}
\bibfield{author}{\bibinfo{person}{Charl Maree}, \bibinfo{person}{Jan~Erik
  Modal}, {and} \bibinfo{person}{Christian~W Omlin}.}
  \bibinfo{year}{2020}\natexlab{}.
\newblock \showarticletitle{Towards responsible AI for financial transactions}.
  In \bibinfo{booktitle}{\emph{2020 IEEE Symposium Series on Computational
  Intelligence (SSCI)}}. IEEE, \bibinfo{pages}{16--21}.
\newblock


\bibitem[McAuley et~al\mbox{.}(2015)]%
        {mcauley2015image}
\bibfield{author}{\bibinfo{person}{Julian McAuley},
  \bibinfo{person}{Christopher Targett}, \bibinfo{person}{Qinfeng Shi}, {and}
  \bibinfo{person}{Anton Van Den~Hengel}.} \bibinfo{year}{2015}\natexlab{}.
\newblock \showarticletitle{Image-based recommendations on styles and
  substitutes}. In \bibinfo{booktitle}{\emph{Proceedings of the 38th
  international ACM SIGIR conference on research and development in information
  retrieval}}. \bibinfo{pages}{43--52}.
\newblock


\bibitem[Newman(2001)]%
        {newman2001clustering}
\bibfield{author}{\bibinfo{person}{Mark~EJ Newman}.}
  \bibinfo{year}{2001}\natexlab{}.
\newblock \showarticletitle{Clustering and preferential attachment in growing
  networks}.
\newblock \bibinfo{journal}{\emph{Physical review E}} \bibinfo{volume}{64},
  \bibinfo{number}{2} (\bibinfo{year}{2001}), \bibinfo{pages}{025102}.
\newblock


\bibitem[Nickel et~al\mbox{.}(2015)]%
        {nickel2015review}
\bibfield{author}{\bibinfo{person}{Maximilian Nickel}, \bibinfo{person}{Kevin
  Murphy}, \bibinfo{person}{Volker Tresp}, {and} \bibinfo{person}{Evgeniy
  Gabrilovich}.} \bibinfo{year}{2015}\natexlab{}.
\newblock \showarticletitle{A review of relational machine learning for
  knowledge graphs}.
\newblock \bibinfo{journal}{\emph{Proc. IEEE}} \bibinfo{volume}{104},
  \bibinfo{number}{1} (\bibinfo{year}{2015}), \bibinfo{pages}{11--33}.
\newblock


\bibitem[Page et~al\mbox{.}(1999)]%
        {page1999pagerank}
\bibfield{author}{\bibinfo{person}{Lawrence Page}, \bibinfo{person}{Sergey
  Brin}, \bibinfo{person}{Rajeev Motwani}, {and} \bibinfo{person}{Terry
  Winograd}.} \bibinfo{year}{1999}\natexlab{}.
\newblock \bibinfo{booktitle}{\emph{The PageRank citation ranking: Bringing
  order to the web.}}
\newblock \bibinfo{type}{{T}echnical {R}eport}. \bibinfo{institution}{Stanford
  InfoLab}.
\newblock


\bibitem[Pan et~al\mbox{.}(2022)]%
        {pan2022neural}
\bibfield{author}{\bibinfo{person}{Liming Pan}, \bibinfo{person}{Cheng Shi},
  {and} \bibinfo{person}{Ivan Dokmani{\'c}}.} \bibinfo{year}{2022}\natexlab{}.
\newblock \showarticletitle{Neural Link Prediction with Walk Pooling}. In
  \bibinfo{booktitle}{\emph{International Conference on Learning
  Representations}}.
\newblock
\urldef\tempurl%
\url{https://openreview.net/forum?id=CCu6RcUMwK0}
\showURL{%
\tempurl}


\bibitem[Qi et~al\mbox{.}(2006)]%
        {qi2006evaluation}
\bibfield{author}{\bibinfo{person}{Yanjun Qi}, \bibinfo{person}{Ziv
  Bar-Joseph}, {and} \bibinfo{person}{Judith Klein-Seetharaman}.}
  \bibinfo{year}{2006}\natexlab{}.
\newblock \showarticletitle{Evaluation of different biological data and
  computational classification methods for use in protein interaction
  prediction}.
\newblock \bibinfo{journal}{\emph{Proteins: Structure, Function, and
  Bioinformatics}} \bibinfo{volume}{63}, \bibinfo{number}{3}
  (\bibinfo{year}{2006}), \bibinfo{pages}{490--500}.
\newblock


\bibitem[Qu et~al\mbox{.}(2021)]%
        {qu2021imgagn}
\bibfield{author}{\bibinfo{person}{Liang Qu}, \bibinfo{person}{Huaisheng Zhu},
  \bibinfo{person}{Ruiqi Zheng}, \bibinfo{person}{Yuhui Shi}, {and}
  \bibinfo{person}{Hongzhi Yin}.} \bibinfo{year}{2021}\natexlab{}.
\newblock \showarticletitle{Imgagn: Imbalanced network embedding via generative
  adversarial graph networks}. In \bibinfo{booktitle}{\emph{Proceedings of the
  27th ACM SIGKDD Conference on Knowledge Discovery \& Data Mining}}.
  \bibinfo{pages}{1390--1398}.
\newblock


\bibitem[Rossi et~al\mbox{.}(2022)]%
        {rossi2022explaining}
\bibfield{author}{\bibinfo{person}{Andrea Rossi}, \bibinfo{person}{Donatella
  Firmani}, \bibinfo{person}{Paolo Merialdo}, {and} \bibinfo{person}{Tommaso
  Teofili}.} \bibinfo{year}{2022}\natexlab{}.
\newblock \showarticletitle{Explaining link prediction systems based on
  knowledge graph embeddings}. In \bibinfo{booktitle}{\emph{Proceedings of the
  2022 International Conference on Management of Data}}.
  \bibinfo{pages}{2062--2075}.
\newblock


\bibitem[Schlichtkrull et~al\mbox{.}(2020)]%
        {schlichtkrull2020interpreting}
\bibfield{author}{\bibinfo{person}{Michael~Sejr Schlichtkrull},
  \bibinfo{person}{Nicola De~Cao}, {and} \bibinfo{person}{Ivan Titov}.}
  \bibinfo{year}{2020}\natexlab{}.
\newblock \showarticletitle{Interpreting graph neural networks for nlp with
  differentiable edge masking}.
\newblock \bibinfo{journal}{\emph{arXiv preprint arXiv:2010.00577}}
  (\bibinfo{year}{2020}).
\newblock


\bibitem[Shchur et~al\mbox{.}(2018)]%
        {shchur2018pitfalls}
\bibfield{author}{\bibinfo{person}{Oleksandr Shchur},
  \bibinfo{person}{Maximilian Mumme}, \bibinfo{person}{Aleksandar Bojchevski},
  {and} \bibinfo{person}{Stephan G{\"u}nnemann}.}
  \bibinfo{year}{2018}\natexlab{}.
\newblock \showarticletitle{Pitfalls of graph neural network evaluation}.
\newblock \bibinfo{journal}{\emph{arXiv preprint arXiv:1811.05868}}
  (\bibinfo{year}{2018}).
\newblock


\bibitem[Tang et~al\mbox{.}(2019)]%
        {tang2019chebnet}
\bibfield{author}{\bibinfo{person}{Shanshan Tang}, \bibinfo{person}{Bo Li},
  {and} \bibinfo{person}{Haijun Yu}.} \bibinfo{year}{2019}\natexlab{}.
\newblock \showarticletitle{ChebNet: Efficient and stable constructions of deep
  neural networks with rectified power units using chebyshev approximations}.
\newblock \bibinfo{journal}{\emph{arXiv preprint arXiv:1911.05467}}
  (\bibinfo{year}{2019}).
\newblock


\bibitem[Veli{\v{c}}kovi{\'c} et~al\mbox{.}(2017)]%
        {velivckovic2017graph}
\bibfield{author}{\bibinfo{person}{Petar Veli{\v{c}}kovi{\'c}},
  \bibinfo{person}{Guillem Cucurull}, \bibinfo{person}{Arantxa Casanova},
  \bibinfo{person}{Adriana Romero}, \bibinfo{person}{Pietro Lio}, {and}
  \bibinfo{person}{Yoshua Bengio}.} \bibinfo{year}{2017}\natexlab{}.
\newblock \showarticletitle{Graph attention networks}.
\newblock \bibinfo{journal}{\emph{arXiv preprint arXiv:1710.10903}}
  (\bibinfo{year}{2017}).
\newblock


\bibitem[Wang et~al\mbox{.}(2020)]%
        {wang2020microsoft}
\bibfield{author}{\bibinfo{person}{Kuansan Wang}, \bibinfo{person}{Zhihong
  Shen}, \bibinfo{person}{Chiyuan Huang}, \bibinfo{person}{Chieh-Han Wu},
  \bibinfo{person}{Yuxiao Dong}, {and} \bibinfo{person}{Anshul Kanakia}.}
  \bibinfo{year}{2020}\natexlab{}.
\newblock \showarticletitle{Microsoft academic graph: When experts are not
  enough}.
\newblock \bibinfo{journal}{\emph{Quantitative Science Studies}}
  \bibinfo{volume}{1}, \bibinfo{number}{1} (\bibinfo{year}{2020}),
  \bibinfo{pages}{396--413}.
\newblock


\bibitem[Wang et~al\mbox{.}(2021)]%
        {wang2021modeling}
\bibfield{author}{\bibinfo{person}{Zhen Wang}, \bibinfo{person}{Bo Zong}, {and}
  \bibinfo{person}{Huan Sun}.} \bibinfo{year}{2021}\natexlab{}.
\newblock \showarticletitle{Modeling Context Pair Interaction for Pairwise
  Tasks on Graphs}. In \bibinfo{booktitle}{\emph{Proceedings of the 14th ACM
  International Conference on Web Search and Data Mining}}.
  \bibinfo{pages}{851--859}.
\newblock


\bibitem[Xiao et~al\mbox{.}(2021)]%
        {xiao2021learning}
\bibfield{author}{\bibinfo{person}{Teng Xiao}, \bibinfo{person}{Zhengyu Chen},
  \bibinfo{person}{Donglin Wang}, {and} \bibinfo{person}{Suhang Wang}.}
  \bibinfo{year}{2021}\natexlab{}.
\newblock \showarticletitle{Learning how to propagate messages in graph neural
  networks}. In \bibinfo{booktitle}{\emph{Proceedings of the 27th ACM SIGKDD
  Conference on Knowledge Discovery \& Data Mining}}.
  \bibinfo{pages}{1894--1903}.
\newblock


\bibitem[Ying et~al\mbox{.}(2019)]%
        {ying2019gnnexplainer}
\bibfield{author}{\bibinfo{person}{Zhitao Ying}, \bibinfo{person}{Dylan
  Bourgeois}, \bibinfo{person}{Jiaxuan You}, \bibinfo{person}{Marinka Zitnik},
  {and} \bibinfo{person}{Jure Leskovec}.} \bibinfo{year}{2019}\natexlab{}.
\newblock \showarticletitle{Gnnexplainer: Generating explanations for graph
  neural networks}.
\newblock \bibinfo{journal}{\emph{Advances in neural information processing
  systems}}  \bibinfo{volume}{32} (\bibinfo{year}{2019}).
\newblock


\bibitem[Yuan et~al\mbox{.}(2020)]%
        {yuan2020xgnn}
\bibfield{author}{\bibinfo{person}{Hao Yuan}, \bibinfo{person}{Jiliang Tang},
  \bibinfo{person}{Xia Hu}, {and} \bibinfo{person}{Shuiwang Ji}.}
  \bibinfo{year}{2020}\natexlab{}.
\newblock \showarticletitle{Xgnn: Towards model-level explanations of graph
  neural networks}. In \bibinfo{booktitle}{\emph{Proceedings of the 26th ACM
  SIGKDD International Conference on Knowledge Discovery \& Data Mining}}.
  \bibinfo{pages}{430--438}.
\newblock


\bibitem[Yuan et~al\mbox{.}(2021)]%
        {yuan2021explainability}
\bibfield{author}{\bibinfo{person}{Hao Yuan}, \bibinfo{person}{Haiyang Yu},
  \bibinfo{person}{Jie Wang}, \bibinfo{person}{Kang Li}, {and}
  \bibinfo{person}{Shuiwang Ji}.} \bibinfo{year}{2021}\natexlab{}.
\newblock \showarticletitle{On explainability of graph neural networks via
  subgraph explorations}. In \bibinfo{booktitle}{\emph{International Conference
  on Machine Learning}}. PMLR, \bibinfo{pages}{12241--12252}.
\newblock


\bibitem[Zhang and Chen(2018)]%
        {zhang2018link}
\bibfield{author}{\bibinfo{person}{Muhan Zhang} {and} \bibinfo{person}{Yixin
  Chen}.} \bibinfo{year}{2018}\natexlab{}.
\newblock \showarticletitle{Link prediction based on graph neural networks}.
\newblock \bibinfo{journal}{\emph{Advances in neural information processing
  systems}}  \bibinfo{volume}{31} (\bibinfo{year}{2018}).
\newblock


\bibitem[Zhang et~al\mbox{.}(2019)]%
        {zhang2019interaction}
\bibfield{author}{\bibinfo{person}{Wen Zhang}, \bibinfo{person}{Bibek Paudel},
  \bibinfo{person}{Wei Zhang}, \bibinfo{person}{Abraham Bernstein}, {and}
  \bibinfo{person}{Huajun Chen}.} \bibinfo{year}{2019}\natexlab{}.
\newblock \showarticletitle{Interaction embeddings for prediction and
  explanation in knowledge graphs}. In \bibinfo{booktitle}{\emph{Proceedings of
  the Twelfth ACM International Conference on Web Search and Data Mining}}.
  \bibinfo{pages}{96--104}.
\newblock


\bibitem[Zhang et~al\mbox{.}(2022)]%
        {zhang2022protgnn}
\bibfield{author}{\bibinfo{person}{Zaixi Zhang}, \bibinfo{person}{Qi Liu},
  \bibinfo{person}{Hao Wang}, \bibinfo{person}{Chengqiang Lu}, {and}
  \bibinfo{person}{Cheekong Lee}.} \bibinfo{year}{2022}\natexlab{}.
\newblock \showarticletitle{Protgnn: Towards self-explaining graph neural
  networks}. In \bibinfo{booktitle}{\emph{Proceedings of the AAAI Conference on
  Artificial Intelligence}}, Vol.~\bibinfo{volume}{36}.
  \bibinfo{pages}{9127--9135}.
\newblock


\bibitem[Zhao et~al\mbox{.}(2022)]%
        {zhao2022exploring}
\bibfield{author}{\bibinfo{person}{Tianxiang Zhao}, \bibinfo{person}{Xiang
  Zhang}, {and} \bibinfo{person}{Suhang Wang}.}
  \bibinfo{year}{2022}\natexlab{}.
\newblock \showarticletitle{Exploring edge disentanglement for node
  classification}. In \bibinfo{booktitle}{\emph{Proceedings of the ACM Web
  Conference 2022}}. \bibinfo{pages}{1028--1036}.
\newblock


\bibitem[Zhou et~al\mbox{.}(2020)]%
        {zhou2020graph}
\bibfield{author}{\bibinfo{person}{Jie Zhou}, \bibinfo{person}{Ganqu Cui},
  \bibinfo{person}{Shengding Hu}, \bibinfo{person}{Zhengyan Zhang},
  \bibinfo{person}{Cheng Yang}, \bibinfo{person}{Zhiyuan Liu},
  \bibinfo{person}{Lifeng Wang}, \bibinfo{person}{Changcheng Li}, {and}
  \bibinfo{person}{Maosong Sun}.} \bibinfo{year}{2020}\natexlab{}.
\newblock \showarticletitle{Graph neural networks: A review of methods and
  applications}.
\newblock \bibinfo{journal}{\emph{AI Open}}  \bibinfo{volume}{1}
  (\bibinfo{year}{2020}), \bibinfo{pages}{57--81}.
\newblock


\end{thebibliography}
